\documentclass{article}

\PassOptionsToPackage{numbers, compress}{natbib}
\PassOptionsToPackage{hyphens}{url}
\usepackage[final]{neurips_mlsb_2025}

\usepackage[utf8]{inputenc} 
\usepackage[T1]{fontenc}    
\usepackage{hyperref}       
\usepackage{url}            
\usepackage{booktabs}       
\usepackage{makecell}
\usepackage{amsfonts}       
\usepackage{nicefrac}       
\usepackage{microtype}      
\usepackage{xcolor}         
\usepackage{amsmath} 
\usepackage{physics}
\usepackage{hyperref}
\usepackage{graphicx}
\usepackage{multirow}
\usepackage{float}
\usepackage[nameinlink]{cleveref} 

\usepackage{pifont}

\newcommand{\cmark}{\ding{51}}%
\newcommand{\xmark}{\ding{55}}%

\usepackage{titlesec}
\titlespacing*{\section}{0pt}{1.5ex plus 1ex minus .2ex}{1ex plus .2ex}
\titlespacing*{\subsection}{0pt}{1.5ex plus 1ex minus .2ex}{1ex plus .2ex}
\usepackage{caption}
\usepackage{titlesec}
\titleformat{\paragraph}[runin]   
  {\normalfont\normalsize\bfseries}
  {}                               
  {0pt}                            
  {}                              
\titlespacing*{\paragraph}{0pt}{1.25ex plus .2ex minus .2ex}{1em}

\usepackage{etoc}

\usepackage{array}

\etocsettagdepth{main}{subsection}
\etocsettagdepth{appendix}{none}
\etocdepthtag.toc{main}

\title{OMTRA: A Multi-Task Generative Model for Structure-Based Drug Design}

\author{%
  Ian Dunn\thanks{Department of Computational and Systems Biology, School of Medicine, University of Pittsburgh} \\
  \And
  Liv Toft\thanks{Ray and Stephanie Lane Computational Biology Department, School of Computer Science, Carnegie Mellon University} \\
  \And
  Tyler Katz\footnotemark[2] \\
  \And
  Juhi Gupta\footnotemark[2] \\
  \And
  Riya Shah\footnotemark[1] \\
  \And
  Ramith Hettiarachchi\footnotemark[2] \\
  \And
  David R. Koes\footnotemark[1] \\
}

\begin{document}

\maketitle

\vspace*{-1em} 

\begin{abstract}
  Structure-based drug design (SBDD) focuses on designing small-molecule ligands that bind to specific protein pockets. Computational methods are integral in modern SBDD workflows and often make use of virtual screening methods via docking or pharmacophore search. Modern generative modeling approaches have focused on improving novel ligand discovery by enabling \textit{de novo} design. In this work, we recognize that these tasks share a common structure and can therefore be represented as different instantiations of a consistent generative modeling framework. We propose a unified approach in OMTRA, a multi-modal flow matching model that flexibly performs many tasks relevant to SBDD, including some with no analogue in conventional workflows. Additionally, we curate a dataset of 500M 3D molecular conformers, complementing protein–ligand data and expanding the chemical diversity available for training. OMTRA obtains state-of-the-art performance on pocket-conditioned \textit{de novo} design and docking; however, the effects of large-scale pretraining and multi-task training are modest. All code, trained models, and dataset for reproducing this work are available at \url{https://github.com/gnina/OMTRA}
\end{abstract}

\section{Introduction}

Deep generative models have rapidly advanced our ability to design and study molecules; this progress will have broad applications. By learning distributions over chemical structures, these models enable tasks ranging from probing biological mechanisms to accelerating therapeutic and general chemical discovery \cite{bilodeau_generative_2022,du_machine_2024,yim_diffusion_2024,morehead_how_2025}.

We focus here on structure-based drug design (SBDD), where the goal is to generate small molecules that bind to a given protein pocket of known structure. Generative approaches have been impactful in two main directions: structure prediction and \textit{de novo} design. In structure prediction, models propose ligand binding poses~\cite{morehead_flowdock_2025,corso_diffdock_2023} or protein–ligand complexes~\cite{abramson_accurate_2024,wohlwend_boltz-1_2025,discovery_chai-1_2024,qiao_state-specific_2024}, improving over classical docking methods by directly sampling plausible structures instead of relying on slow search algorithms. In \textit{de novo} design, generative models sample the identity of the molecules themselves, sometimes in addition to the 3D structure. De novo generative models can also be conditioned on desired properties such as a binding partner \cite{ragoza2022generating,dunn_flowmol3_2025,campbell_generative_2024,cremer_flowr_2025,joshi_all-atom_2025,jing_generating_2025,qin_defog_2025,chen_target_2025}.

Current state of the art methods for both structure prediction and \textit{de novo} design are primarily \emph{transport-based generative models}—a broad class that includes diffusion~\cite{sohl-dickstein_deep_2015,ho_denoising_2020,song_score-based_2021}, flow matching~\cite{lipman_flow_2023,tong_improving_2023,liu_flow_2022}, and stochastic interpolant models~\cite{albergo_stochastic_2023}. These methods also primarily use attributed point-cloud or sequence-based representations of molecules. These approaches treat molecular systems as objects composed of distinct modalities, and may therefore be additionally characterized as \emph{multi-modal transport-based models}. Atoms have both continuous positions and discrete identities~\cite{dunn_flowmol3_2025,joshi_all-atom_2025,irwin_efficient_2024,nikitin_geom-drugs_2025}. Protein residues are often described similarly but their structure is sometimes decomposed into a product space of rotations, translations, and/or torsional angles~\cite{schneuing_multi-domain_2025,huguet_sequence-augmented_2024,jumper_highly_2021,ingraham_illuminating_2023}. These methods simultaneously apply transport-based modeling to each modality. To our knowledge, this unifying perspective has not been made explicit. After recognizing that many applications or generative tasks may be formulated under one framework, a natural question therefore arises: could one model be trained for many tasks, and could this yield improved performance from transfer learning?

Building on this idea, we introduce \textbf{OMTRA}, a multi-task generative model for SBDD. OMTRA supports ligands, protein pockets, and pharmacophores, co-factors, ions, and post-translation modifications. Tasks can be defined by dividing modalities into those generated, held fixed as conditioning information, and those absent. OMTRA is capable of handling arbitrary modality partitions and therefore can perform molecular docking, de novo design, conformer generation, pharmacophore-conditioned design, and related tasks. OMTRA's architecture enables parameter sharing across tasks and learning from disparate data sources.

Finally, we contribute a large-scale dataset of 500M 3D molecular conformers, constructed from public chemical libraries and released in a deep learning–ready format. This dataset expands chemical diversity available for training and complements protein–ligand data, facilitating multi-task learning across heterogeneous molecular modalities.

\begin{figure}[t]
    \centering
    \includegraphics[width=0.99\linewidth]{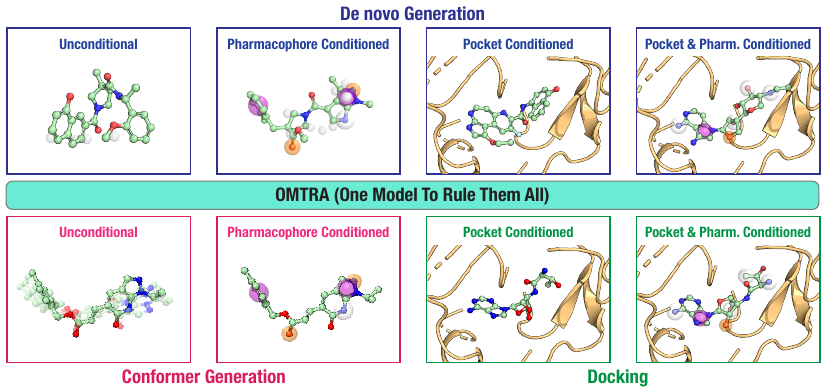}
    \caption{\textbf{OMTRA: A Flexible Multi-Task Generative Model for Structure-Based Drug Design.} OMTRA is capable of performing \textit{de novo} design, conformer generation and docking. It can be guided by conditioning on structural information, such as a protein pocket and pharmacophores.}
    \label{fig:omtra_main}
\end{figure}

\section{Methods}

\subsection{Flow Matching}  \label{sec:fm-intro}

Flow matching \cite{tong_improving_2023,albergo_stochastic_2023,lipman_flow_2023,liu_flow_2022} prescribes a method to interpolate between two probability distributions. This is accomplished modeling a family of intermediate distributions $\{p_t\}_{t\in[0,1]}$ with $p_0 = q_{\text{source}}$ and $p_1 = q_{\text{target}}$. The general strategy is to draw samples from a tractable $q_{\text{source}}$ and use a learned process $p_t$ to transport them to samples from an otherwise intractable $q_{\text{target}}$. For continuous variables $x \in \mathbb{R}^d$, the marginal process is sampled by a vector field defining an ordinary differential equation $u_t(x)=\tfrac{dx}{dt}$. For discrete data $x \in \mathcal{V}^N$ (sequences of tokens from a vocabulary $\mathcal{V}$), trajectories are governed by continuous-time Markov chains (CTMCs)~\cite{campbell_generative_2024,gat_discrete_2024}. In both cases the sampling process can be parameterized by a neural network trained to approximate the final system state given the current state $\mathbb{E}[x_1|x_t]$.

Multi-modal flow matching is a powerful extension of this framework wherein complex objects are modeled as collections of distinct dependent variables, referred to as modalities. For example, atom positions and atom types form two modalities for which joint sampling can enable \textit{de novo} molecule generation. Multi-modal flow matching \cite{dunn_flowmol3_2025, campbell_generative_2024} extends the base flow matching framework to sample such data structures. One neural network is trained with multiple prediction heads to minimize a weighted sum of flow matching losses on each modality. Modalities in the system are generated by simultaneous sampling of an ODE or CTMC on each modality. A full description of our flow matching formulation is provided in Appendix \ref{ap:flow-matching}.

\subsection{Problem Setup}

We represent a biomolecular system cropped to 8\AA\ around a ground-truth ligand as a heterogeneous graph 
$G = (V, E)$ with node types 
$\mathcal{T}_V = \{\text{ligand atom}, \text{protein atom}, \text{pharmacophore}, \text{other}\}$ 
and edge types $\mathcal{T}_E$ (e.g., ligand--ligand, ligand--protein).  
Each node or edge carries one or more \emph{modalities}, discrete (e.g., atom type, bond order) or continuous (e.g., 3D coordinates), 
so that $\mathcal{M} = \mathcal{M}_{\text{disc}} \cup \mathcal{M}_{\text{cont}}$. A \emph{task} $\tau$ specifies a partition of modalities $\mathcal{M} = \mathcal{M}_{\text{gen}}^\tau \cup \mathcal{M}_{\text{cond}}^\tau \cup \mathcal{M}_{\text{abs}}^\tau$ referred to as generated, conditioning, and absent modalities, respectively. Generated modalities, $\mathcal{M}_{\text{gen}}^\tau$ are sampled by a flow matching model. 

An instantiation of OMTRA requires choosing a set of tasks $\mathcal{T} = \{\tau_1, \dots, \tau_K\}$ to support. If OMTRA is instantiated as a multi-task model (i.e., $|\mathcal{T}| > 1$), weights are shared across tasks. See Appendix~\ref{ap:problem-setup} for a full formalization. 

A full list of modalities and tasks currently supported, as well as tasks we plan to support in the near feature, are provided in Appendix \ref{ap:tasks-and-modalities}.

\subsection{Architecture} \label{sec:arch}

OMTRA extends FlowMol3 \cite{dunn_flowmol3_2025}, a geometric graph neural network, to heterogeneous graphs with type-specific convolutions.  
Nodes $i \in V$ carry Cartesian positions $x_i \in \mathbb{R}^3$, scalar features $s_i \in \mathbb{R}^{d_s}$, and vector features $v_i \in \mathbb{R}^{d_v \times 3}$.  
Edges $(i,j) \in E$ are typed; ligand--ligand and other--other edges additionally carry scalar features $e_{ij} \in \mathbb{R}^{d_e}$.  
Operations on $(x_i, v_i)$ are SE(3)-equivariant, while operations on $(s_i, e_{ij})$ are SE(3)-invariant. Node vectors $v_i$ are first-order geometric vectors relative to $x_i$. Equivariance is enabled by a variant of GVPs~\cite{jing_equivariant_2021} introduced in \citet{dunn_flowmol3_2025}.

A \emph{graph convolution} consists of edge-type–specific message functions $\phi_r$ that generate messages $m_{i \leftarrow j}^r$ based on $(s_i, s_j, v_i, v_j, e_{ij}, x_i - x_j)$, followed by node-type–specific update functions $\psi_\alpha$ that integrate the aggregated messages $M_i$ to update $(s_i, v_i)$.  After 2 graph convolutions, positions and edge features are updated via node-wise GVPs and edge-wise MLPs. OMTRA repeats this sequence (called a \emph{ConvolutionBlock}) 4 times. For modalities defining positions, the final $x_i$ are taken directly from the last \emph{ConvolutionBlock}. Latent node and edge features are mapped to categorical logits by shallow MLP heads. The architecture is described fully in Appendix \ref{ap:arch}.

\section{Experiments}

\subsection{Datasets}

To enable pretraining on protein-free tasks, we curated the Pharmit Dataset containing 500M 3D ligands and their pharmacophores. The Pharmit Dataset is further described in Appendix \ref{ap:pharmit}.

For ablation studies we train on protein-ligand complexes from the Plinder dataset~\cite{plinder} using their rigorously designed splits. For fair evaluation of pocket-conditioned \textit{de novo} design, we also train and evaluate OMTRA on the Crossdocked dataset \cite{francoeur_three-dimensional_2020}, using the train/test splits from \citet{luo_3d_2022}. These splits are less rigorous than those of the original Crossdocked authors and likely contain information leakage, but they are widely reused and thus allow direct comparison to prior work. For comparing OMTRA to other docking methods, we use the PoseBusters Benchmark set \cite{buttenschoen_posebusters_2024}. This is a moderately difficult docking benchmark comprising 428 diverse protein-ligand complexes. All datasets are described in detail in Appendix \ref{ap:datasets}. We did not take care to remove overlapping examples from the Plinder training set before evaluating on the Posebusters Benchmark Set, and as such our estimates of docking performance on PoseBusters may be optimistic.

\subsection{Evaluation and Baselines} \label{sec:baselines}

We evaluate OMTRA on pocket-conditioned \textit{de novo} design, docking, and pharmacophore conditioning tasks, comparing against established baselines under a unified protocol.  

\paragraph{Pocket-Conditioned De Novo Design}
Generated ligands are assessed with PoseBusters~\cite{buttenschoen_posebusters_2024} for chemical/geometric plausibility. The fraction of ligands passing all relevant checks in the PoseBusters suite is reported as \%PB-Valid. Strain energy ($E_{strain}$) and protein-ligand interactions are measured using PoseCheck~\cite{harris_benchmarking_2023}. Posecheck reports the number of hydrogen bond donors, acceptors, and hydrophobic groups on the ligand that are interacting with the protein, normalized by the number of atoms in the ligand. We report ``\% Interaction Parity'' defined as the fraction of generated ligands that have normalized interaction counts equal to or greater than that of their ground-truth ligand. Additionally we report the fraction of ligands that achieve interaction parity and are PB-valid as ``\% PB-Valid+IP''.  We compare OMTRA to DrugFlow~\cite{schneuing_multi-domain_2025}, DiffSBDD~\cite{schneuing_structure-based_2024}, TargetDiff~\cite{guan_3d_2023}, and Pocket2Mol~\cite{peng_pocket2mol_2025}, using 100 test pockets with 100 ligands per pocket. Full experimental details of our evaluation methods are in Appendices \ref{ap:denovo-eval}.

\paragraph{Docking}
Docked poses are evaluated with PoseBusters in docking mode, reporting physical plausibility and RMSD $<2$\AA\ to the ground truth, along with Top-$N$ success rates. To perform a Top-$N$ analysis on OMTRA, samples are ranked by the Vina scoring function~\cite{eberhardt_autodock_2021} combined with a penalty for docked poses that violate that don't preserve the chirality of the input ligand.  OMTRA is trained on Plinder~\cite{plinder} and evaluated on the PoseBusters Benchmark, sampling 40 poses per receptor. Baseline docking methods include AlphaFold3~\cite{abramson_accurate_2024}, SurfDock~\cite{cao_surfdock_2025}, Gnina~\cite{mcnutt_Gnina_2025}, AutoDock Vina~\cite{eberhardt_autodock_2021}. Performance is reported as the fraction of systems with at least one PoseBusters-valid pose among the top-$N$ predictions (N=1, 5). Full experimental details of our evaluation methods are in Appendices \ref{ap:docking-eval}.

\paragraph{Pharamcophore Conditioning} When pharmacophore centers are provided as conditioning information, we report ``\% Pharm Matching'' which is the fraction of pharmacophore centers in the input conditioning information that are ``satisfied'' by the generated/docked ligand. A pharmacophore condition is satisfied if the generated ligand has a pharmacophore center of the same type within 1\AA\ of the conditioning pharmacophore. We also report ``interaction recovery'' which is fraction of ligands that make all of the same types of interactions with the same protein residues as the ground-truth ligand. Pharmacophores are described further in Appendix \ref{ap:pharms}.

\subsection{Ablations} \label{sec:ablations}

We train multiple variants on OMTRA to investigate the effects of three features. \textbf{Ligand-only pretraining:} Single-task OMTRA models are trained for either \textit{de novo} design or docking, with and without protein-free pretraining on the Pharmit dataset.  
\textbf{Multi-task training:} OMTRA is trained jointly on \textit{de novo} design and docking, initialized from Pharmit-pretrained models, and compared to single-task counterparts. For all ablation studies, we train and evaluate on the Plinder dataset. We attempt to match the amount of compute spent on each task between single- and multi- task models. Single-task models are trained for 200k steps and multi-task models are trained for 400k steps, with an even balance between \textit{de novo} design and docking. \textbf{Protein + pharmacophore conditioning:} OMTRA is trained to simultaneously use protein and pharmacophore conditioning. This variant can perform design and docking guided by user-supplied, interpretable priors over protein-ligand interaction.

\section{Results}

\paragraph{OMTRA Performance on \textit{De Novo} Design} Table \ref{tab:crossdocked-denovo} shows that OMTRA surpasses or matches existing \textit{de novo} generative models across all evaluations performed. By building on FlowMol3~\cite{dunn_flowmol3_2025}, a state-of-the-art model for unconditional \textit{de novo} molecule design, OMTRA is able to obtain the highest physical plausibility of methods evaluated ($\approx 90\%$ PB-Valid). The mean strain energy and Vina scores are practically close to that of the training data. OMTRA produces ligands that are both physically plausible and produce native-like interactions with the protein 37\% more frequently than the next best method (26 vs 19 \%PB-Valid+IP). Although OMTRA improves upon existing \textit{de novo} generative models, there remain practically significant differences between real and the average generated molecule. Most notably, only 26\% of samples are both PB-valid and achieve native interaction parity.

\begin{table}
\centering
\small
\caption{\textbf{Pocket-Conditioned De Novo Design}. A Multi-Task OMTRA model is compared to existing generative models on the Crossdocked dataset. Metrics are averaged over 100 pockets, with 100 samples drawn per pocket. PB-Valid+IP is the percent of ligands that are both PB-valid and achieve interaction parity. Evaluation details are provided in Section \ref{sec:baselines}}
\label{tab:crossdocked-denovo}
\resizebox{0.85\textwidth}{!}{%
\begin{tabular}{lrrrrr}
\toprule
 & \makecell{PB-Valid\\(\%)($\uparrow$)}
 & \makecell{$E_{strain}$\\(kcal/mol)($\downarrow$)}
 & \makecell{Vina\\(kcal/mol)($\downarrow$)} 
 & \makecell{Interaction\\Parity\\(\%) ($\uparrow$)} 
 & \makecell{PB-Valid+IP\\(\%) ($\uparrow$)} \\
\midrule
Dataset & 97.7 & 21.3 & -7.1 & 100 & 97.7 \\
OMTRA & \textbf{89.8} & 29.5 & \textbf{-6.8} & \textbf{27} & \textbf{26} \\
Pocket2Mol & 86.6 & \textbf{24.0} & -4.5 & 22 & 19 \\
DrugFlow & 73.4 & 72.5 & -5.9 & 21 & 17 \\
TargetDiff & 47.7 & 446 & \textbf{-6.8} & 20 & 11 \\
DiffSBDD & 38.2 & 546 & -2.8 & 18 & 7 \\
\bottomrule
\end{tabular}
}
\end{table}


\paragraph{OMTRA achieves SOTA re-docking performance} Re-docking results on the PoseBusters Benchmark Set in Table \ref{tab:pbdock} show that OMTRA achieves highly accurate re-docking. After sampling and ranking 40 poses for a pocket, 91\% of the top-ranked poses are PB-valid; that is they are within 2\AA\ RMSD of the ground-truth ligand and satisfy a suite of additional physical plausibility checks. OMTRA's top-1 docking accuracy exceeds that of all other models evaluated including AlphaFold3 with pocket residues specified~\cite{abramson_accurate_2024}. Additional results (Figure \ref{fig:posebusters_combined}) show that OMTRA-docked poses are not PB-valid most often due to unconserved tetrahedral chirality. We alleviate this issue by including a penalty for incorrect chirality in our ranking algorithm as done in \citet{abramson_accurate_2024}. Methods for enforcing chirality preservation \cite{wohlwend_boltz-1_2025,nikitin_geom-drugs_2025,anishchenko_modeling_2025} exist and would likely further enhance docking performance.

\begin{table}
\caption{\textbf{Docking Success Rates on the PoseBusters Benchmark Set}. AlphaFold3~\cite{abramson_accurate_2024} and Surfdock~\cite{cao_surfdock_2025} results are taken directly from their respective publications. Vina results are taken from \citet{buttenschoen_posebusters_2024}. Additional details in Section \ref{sec:baselines}.}
\label{tab:pbdock}
\centering
\begin{tabular}{lrr|rr}
\toprule
 & \multicolumn{2}{c}{$N=1$} & \multicolumn{2}{c}{$N=5$} \\
 & \% $\le$ 2\AA\ & \%PB-Valid & \% $\le$ 2\AA\ & \%PB-Valid \\
\midrule
OMTRA & 92 & \textbf{91} & \textbf{97} & \textbf{96} \\
AlphaFold3 & \textbf{93} & 84 & - & - \\
SurfDock & 78 & 40 & 81 & 79 \\
Gnina & 65 & 64 & 82 & 81 \\
Vina & 60 & 58 & - & - \\
\bottomrule
\end{tabular}
\end{table}


\paragraph{Effects of Pretraining and Multi-Task Training on Interaction and Plausibility} Table \ref{tab:exp1} suggests that the effects of pretraining and multi-task training are mostly modest and inconsistent across tasks. Adding protein-free pretraining improves physical plausibility and interaction parity (+9 and +2 percentage points, respectively) but reduces the PB-validity in docking by 4 percentage points. In contrast, the addition of multi-task training impairs \textit{de novo} design while slightly improving docking performance.

\begin{table}
\caption{\textbf{Effect of Pretraining and Multi-Task Training on \textit{De Novo} Design and Docking}. All models are trained and evaluated on Plinder~\cite{plinder} train/test splits. The top row contains metric values computed on the ground-truth protein-ligand systems where applicable. Docking success rates are computed from the top-5 poses per system. OMTRA variations are described in Section \ref{sec:ablations} }
\label{tab:exp1}
\resizebox{\textwidth}{!}{%
\begin{tabular}{cc|rrrr|rr}
\toprule
  &  & \multicolumn{4}{c|}{\textit{De Novo} Design} & \multicolumn{2}{c}{Docking} \\
 Multitask & Pretrained & \% PB-Valid & $E_{strain}$ & \% IP & \% PB-Valid + IP & \% RMSD $\le$ 2\AA\ & \% PB-Valid \\
\midrule
& & 94.0 & 38.1 & 100 & 94.0 &  &  \\
\xmark & \xmark & 60.8 & 91.0 & 13.0 & 10.0 & 96.9 & 92.7 \\
\xmark & \cmark  & 69.1 & 65.3 & 15.0 & 12.0 & 92.9 & 88.8 \\
\cmark  & \cmark  & 67.0 & 71.4 & 9.0 & 7.0 & 94.8 & 91.7 \\
\bottomrule
\end{tabular}
}
\end{table}

\paragraph{Pharmacophore conditioning enables enhanced interaction design} The impact of pharmacophore-conditioning on \textit{de novo} ligand design and docking is quantified in Table \ref{tab:exp3}. When provided with pharmacophore conditioning, OMTRA's ability to design protein-ligand interactions is substantially enhanced, indicated by a 10 percentage point increase in interaction recovery for \textit{de novo} design. Similarly, pharmacophore conditioning significantly enhances the quality of docked poses. Additional experiments (Section \ref{ap:pharm-effect}) suggest that for \textit{de novo} design, adding 3 pharmacophores reduces the amount of sampling required by 56\%. Users with prior knowledge about their target can obtain more accurate predictions from OMTRA with less sampling.

\begin{table}
\caption{\textbf{Effect of Pharmacophore Conditioning on Pocket-Conditioned Design and Docking}. Evaluating one OMTRA model on four tasks: pocket-conditioned \textit{de novo} design and docking, with and without additional pharmacophore conditioning. Evaluation is performed on 100 systems from the Plinder test set having $>20$ ligand heavy atoms, drawing 100 samples per system. Docking success rates are for the top-5 ligands in each system as ranked by their Vina score.}
\centering
\small
\label{tab:exp3}
\begin{tabular}{llrr}
\toprule
 & & Prot Conditioning & Prot + Pharm Conditioning \\
\midrule
\multirow[t]{3}{*}{denovo design} & \% PB-Valid & 67.0 & 70.3 \\
 & interaction recovery & 50.8 & 60.7 \\
 & \% Pharm Matches & - & 97.2 \\
\cmidrule(lr{0.25em}){1-4}
 \multirow[t]{5}{*}{docking} & \% RMSD $\le$ 2\AA\ & 94.8 & 99.0 \\
 & \% PB-Valid & 91.7 & 92.9 \\
 & interaction recovery & 80.6 & 90.7 \\
 & \% Pharm Matches & - & 99.5 \\
\cline{1-4}
\bottomrule
\end{tabular}
\end{table}

\section{Conclusion}

OMTRA is a flexible multi-task generative model for structure-based drug design, achieving state-of-the-art performance in both \textit{de novo} design and molecular docking. It uniquely supports simultaneous pocket and pharmacophore conditioning, enabling the incorporation of user-provided knowledge of protein–ligand interactions to improve accuracy and reduce the amount of sampling necessary.

Our experiments indicate that the benefits of large-scale, protein-free pretraining and multi-task training are modest and not uniformly positive. OMTRA can be trained on multiple tasks while matching the performance of single-task models trained in isolation. Despite a field-wide shift toward multi-task molecular generative models, empirical evidence for meaningful transfer learning remains limited. In our view, whether molecular generative models can reliably leverage transfer across tasks is still an open question. Future work will use OMTRA as a testbed for investigating architectural choices and generative modeling frameworks that may confer this capability.

The current OMTRA architecture represents a minimal extension of an unconditional molecular generative model to a multi-task, heterogeneous-graph setting, leaving substantial room for improvement. Notably, OMTRA can be extended to generate protein structure as an additional modality. This capability would support tasks involving flexible or unknown protein conformations; use-cases in which no holo structure is available and which represent a more realistic and challenging design scenario than the ones considered in our present evaluations. OMTRA’s flexible multi-modal formulation thus makes it not only a versatile tool, but also a platform for advancing molecular generative modeling.

Alongside OMTRA, we release the Pharmit Dataset: one of the largest open-source collections of 3D molecules to date. The Pharmit Dataset is stored in a scalable format with a accessible API for efficient programmatic access so that it can readily incorporated into large-scale machine learning applications. All code, trained models, and dataset for reproducing this work are available at \url{https://github.com/gnina/OMTRA}.

\clearpage
\bibliography{tyler_references,omtra,liv_references,juhi_references,newbib}

@misc{luo_crossdocked_external_splits,
	title = {A 3D Generative Model for Structure-Based Drug Design},
	url = {https://arxiv.org/abs/2203.10446},
	doi = {10.48550/arXiv.2203.10446},
	urldate = {2025-09-21},
	publisher = {arXiv},
	author = {Luo, Shitong and Guan, Jiaqi and Ma, Jianzhu and Peng, Jian},
	month = nov,
	year = {2022},
    note = {arXiv:2203.10446 [q-bio]},
}

@article{liu_pdbbind_2017,
	title = {Forging the {Basis} for {Developing} {Protein}–{Ligand} {Interaction} {Scoring} {Functions}},
	volume = {50},
	issn = {0001-4842},
	url = {https://doi.org/10.1021/acs.accounts.6b00491},
	doi = {10.1021/acs.accounts.6b00491},
	number = {2},
	urldate = {2025-12-04},
	journal = {Accounts of Chemical Research},
	author = {Liu, Zhihai and Su, Minyi and Han, Li and Liu, Jie and Yang, Qifan and Li, Yan and Wang, Renxiao},
	month = feb,
	year = {2017},
	note = {Publisher: American Chemical Society},
	pages = {302--309},
}

@misc{schneuing_structure-based_2024,
	title = {Structure-based Drug Design with Equivariant Diffusion Models},
	url = {http://arxiv.org/abs/2210.13695},
	doi = {10.48550/arXiv.2210.13695},
	number = {{arXiv}:2210.13695},
	publisher = {{arXiv}},
	author = {Schneuing, Arne and Harris, Charles and Du, Yuanqi and Didi, Kieran and Jamasb, Arian and Igashov, Ilia and Du, Weitao and Gomes, Carla and Blundell, Tom and Lio, Pietro and Welling, Max and Bronstein, Michael and Correia, Bruno},
	urldate = {2025-09-19},
	date = {2024-09-23},
	langid = {english},
	eprinttype = {arxiv},
	eprint = {2210.13695 [q-bio]},
}

@misc{schneuing_multi-domain_2025,
      title={Multi-domain Distribution Learning for De Novo Drug Design}, 
      author={Arne Schneuing and Ilia Igashov and Adrian W. Dobbelstein and Thomas Castiglione and Michael Bronstein and Bruno Correia},
      year={2025},
      eprint={2508.17815},
      archivePrefix={arXiv},
      primaryClass={cs.LG},
      url={https://arxiv.org/abs/2508.17815}, 
}

@misc{guan_3d_2023,
	title = {3D Equivariant Diffusion for Target-Aware Molecule Generation and Affinity Prediction},
	url = {http://arxiv.org/abs/2303.03543},
	doi = {10.48550/arXiv.2303.03543},
	number = {{arXiv}:2303.03543},
	publisher = {{arXiv}},
	author = {Guan, Jiaqi and Qian, Wesley Wei and Peng, Xingang and Su, Yufeng and Peng, Jian and Ma, Jianzhu},
	urldate = {2025-09-19},
	date = {2023-03-06},
	langid = {english},
	eprinttype = {arxiv},
	eprint = {2303.03543 [q-bio]},
    year={2023},
}

@article{ragoza2022generating,
  title={Generating 3D molecules conditional on receptor binding sites with deep generative models},
  author={Ragoza, Matthew and Masuda, Tomohide and Koes, David Ryan},
  journal={Chemical science},
  volume={13},
  number={9},
  pages={2701--2713},
  year={2022},
  publisher={Royal Society of Chemistry}
}

@misc{tong_improving_2023,
	title = {Improving and generalizing flow-based generative models with minibatch optimal transport},
	url = {http://arxiv.org/abs/2302.00482},
	doi = {10.48550/arXiv.2302.00482},
	urldate = {2023-09-20},
	publisher = {arXiv},
	author = {Tong, Alexander and Malkin, Nikolay and Huguet, Guillaume and Zhang, Yanlei and Rector-Brooks, Jarrid and Fatras, Kilian and Wolf, Guy and Bengio, Yoshua},
	month = jul,
	year = {2023},
	note = {arXiv:2302.00482 [cs]},
}

@misc{sohl-dickstein_deep_2015,
	title = {Deep {Unsupervised} {Learning} using {Nonequilibrium} {Thermodynamics}},
	url = {http://arxiv.org/abs/1503.03585},
	doi = {10.48550/arXiv.1503.03585},
	urldate = {2024-03-14},
	publisher = {arXiv},
	author = {Sohl-Dickstein, Jascha and Weiss, Eric A. and Maheswaranathan, Niru and Ganguli, Surya},
	month = nov,
	year = {2015},
	note = {arXiv:1503.03585 [cond-mat, q-bio, stat]},
}

@misc{ho_denoising_2020,
	title = {Denoising {Diffusion} {Probabilistic} {Models}},
	url = {http://arxiv.org/abs/2006.11239},
	doi = {10.48550/arXiv.2006.11239},
	urldate = {2024-03-14},
	publisher = {arXiv},
	author = {Ho, Jonathan and Jain, Ajay and Abbeel, Pieter},
	month = dec,
	year = {2020},
	note = {arXiv:2006.11239 [cs, stat]},
}

@misc{le_navigating_2023,
	title = {Navigating the {Design} {Space} of {Equivariant} {Diffusion}-{Based} {Generative} {Models} for {De} {Novo} {3D} {Molecule} {Generation}},
	url = {http://arxiv.org/abs/2309.17296},
	doi = {10.48550/arXiv.2309.17296},
	urldate = {2024-03-14},
	publisher = {arXiv},
	author = {Le, Tuan and Cremer, Julian and Noé, Frank and Clevert, Djork-Arné and Schütt, Kristof},
	month = nov,
	year = {2023},
	note = {arXiv:2309.17296 [cs]},
}

@misc{albergo_stochastic_2023,
	title = {Stochastic {Interpolants}: {A} {Unifying} {Framework} for {Flows} and {Diffusions}},
	shorttitle = {Stochastic {Interpolants}},
	url = {http://arxiv.org/abs/2303.08797},
	doi = {10.48550/arXiv.2303.08797},
	urldate = {2024-03-18},
	publisher = {arXiv},
	author = {Albergo, Michael S. and Boffi, Nicholas M. and Vanden-Eijnden, Eric},
	month = nov,
	year = {2023},
	note = {arXiv:2303.08797 [cond-mat]},
}

@misc{jing_equivariant_2021,
	title = {Equivariant {Graph} {Neural} {Networks} for {3D} {Macromolecular} {Structure}},
	url = {http://arxiv.org/abs/2106.03843},
	doi = {10.48550/arXiv.2106.03843},
	urldate = {2024-03-20},
	publisher = {arXiv},
	author = {Jing, Bowen and Eismann, Stephan and Soni, Pratham N. and Dror, Ron O.},
	month = jul,
	year = {2021},
	note = {arXiv:2106.03843 [cs, q-bio]},
}

@misc{lipman_flow_2023,
	title = {Flow {Matching} for {Generative} {Modeling}},
	url = {http://arxiv.org/abs/2210.02747},
	doi = {10.48550/arXiv.2210.02747},
	urldate = {2024-03-30},
	publisher = {arXiv},
	author = {Lipman, Yaron and Chen, Ricky T. Q. and Ben-Hamu, Heli and Nickel, Maximilian and Le, Matt},
	month = feb,
	year = {2023},
	note = {arXiv:2210.02747 [cs, stat]},
}

@misc{song_score-based_2021,
	title = {Score-{Based} {Generative} {Modeling} through {Stochastic} {Differential} {Equations}},
	url = {http://arxiv.org/abs/2011.13456},
	doi = {10.48550/arXiv.2011.13456},
	urldate = {2024-04-02},
	publisher = {arXiv},
	author = {Song, Yang and Sohl-Dickstein, Jascha and Kingma, Diederik P. and Kumar, Abhishek and Ermon, Stefano and Poole, Ben},
	month = feb,
	year = {2021},
	note = {arXiv:2011.13456 [cs, stat]},
}

@misc{liu_flow_2022,
	title = {Flow {Straight} and {Fast}: {Learning} to {Generate} and {Transfer} {Data} with {Rectified} {Flow}},
	shorttitle = {Flow {Straight} and {Fast}},
	url = {http://arxiv.org/abs/2209.03003},
	doi = {10.48550/arXiv.2209.03003},
	urldate = {2024-04-04},
	publisher = {arXiv},
	author = {Liu, Xingchao and Gong, Chengyue and Liu, Qiang},
	month = sep,
	year = {2022},
	note = {arXiv:2209.03003 [cs]},
}

@article{ingraham_illuminating_2023,
	title = {Illuminating protein space with a programmable generative model},
	volume = {623},
	copyright = {2023 The Author(s)},
	issn = {1476-4687},
	url = {https://www.nature.com/articles/s41586-023-06728-8},
	doi = {10.1038/s41586-023-06728-8},
	language = {en},
	number = {7989},
	urldate = {2024-04-24},
	journal = {Nature},
	author = {Ingraham, John B. and Baranov, Max and Costello, Zak and Barber, Karl W. and Wang, Wujie and Ismail, Ahmed and Frappier, Vincent and Lord, Dana M. and Ng-Thow-Hing, Christopher and Van Vlack, Erik R. and Tie, Shan and Xue, Vincent and Cowles, Sarah C. and Leung, Alan and Rodrigues, João V. and Morales-Perez, Claudio L. and Ayoub, Alex M. and Green, Robin and Puentes, Katherine and Oplinger, Frank and Panwar, Nishant V. and Obermeyer, Fritz and Root, Adam R. and Beam, Andrew L. and Poelwijk, Frank J. and Grigoryan, Gevorg},
	month = nov,
	year = {2023},
	note = {Publisher: Nature Publishing Group},
	pages = {1070--1078},
}

@misc{campbell_generative_2024,
	title = {Generative {Flows} on {Discrete} {State}-{Spaces}: {Enabling} {Multimodal} {Flows} with {Applications} to {Protein} {Co}-{Design}},
	shorttitle = {Generative {Flows} on {Discrete} {State}-{Spaces}},
	url = {http://arxiv.org/abs/2402.04997},
	doi = {10.48550/arXiv.2402.04997},
	urldate = {2024-09-17},
	publisher = {arXiv},
	author = {Campbell, Andrew and Yim, Jason and Barzilay, Regina and Rainforth, Tom and Jaakkola, Tommi},
	month = jun,
	year = {2024},
	note = {arXiv:2402.04997 [cs, q-bio, stat]},
}

@misc{gat_discrete_2024,
	title = {Discrete {Flow} {Matching}},
	url = {http://arxiv.org/abs/2407.15595},
	doi = {10.48550/arXiv.2407.15595},
	urldate = {2024-09-17},
	publisher = {arXiv},
	author = {Gat, Itai and Remez, Tal and Shaul, Neta and Kreuk, Felix and Chen, Ricky T. Q. and Synnaeve, Gabriel and Adi, Yossi and Lipman, Yaron},
	month = jul,
	year = {2024},
	note = {arXiv:2407.15595 [cs]},
}

@misc{irwin_efficient_2024,
	title = {Efficient {3D} {Molecular} {Generation} with {Flow} {Matching} and {Scale} {Optimal} {Transport}},
	url = {http://arxiv.org/abs/2406.07266},
	doi = {10.48550/arXiv.2406.07266},
	urldate = {2024-09-17},
	publisher = {arXiv},
	author = {Irwin, Ross and Tibo, Alessandro and Janet, Jon Paul and Olsson, Simon},
	month = jun,
	year = {2024},
	note = {arXiv:2406.07266 [cs]},
}

@misc{huguet_sequence-augmented_2024,
	title = {Sequence-{Augmented} {SE}(3)-{Flow} {Matching} {For} {Conditional} {Protein} {Backbone} {Generation}},
	url = {http://arxiv.org/abs/2405.20313},
	doi = {10.48550/arXiv.2405.20313},
	urldate = {2024-09-21},
	publisher = {arXiv},
	author = {Huguet, Guillaume and Vuckovic, James and Fatras, Kilian and Thibodeau-Laufer, Eric and Lemos, Pablo and Islam, Riashat and Liu, Cheng-Hao and Rector-Brooks, Jarrid and Akhound-Sadegh, Tara and Bronstein, Michael and Tong, Alexander and Bose, Avishek Joey},
	month = may,
	year = {2024},
	note = {arXiv:2405.20313 [cs, q-bio]},
}

@misc{reidenbach_applications_2025,
	title = {Applications of {Modular} {Co}-{Design} for {De} {Novo} {3D} {Molecule} {Generation}},
	url = {http://arxiv.org/abs/2505.18392},
	doi = {10.48550/arXiv.2505.18392},
	urldate = {2025-07-21},
	publisher = {arXiv},
	author = {Reidenbach, Danny and Nikitin, Filipp and Isayev, Olexandr and Paliwal, Saee},
	month = may,
	year = {2025},
	note = {arXiv:2505.18392 [cs]},
}

@misc{nikitin_geom-drugs_2025,
	title = {{GEOM}-{Drugs} {Revisited}: {Toward} {More} {Chemically} {Accurate} {Benchmarks} for {3D} {Molecule} {Generation}},
	shorttitle = {{GEOM}-{Drugs} {Revisited}},
	url = {http://arxiv.org/abs/2505.00169},
	doi = {10.48550/arXiv.2505.00169},
	urldate = {2025-07-24},
	publisher = {arXiv},
	author = {Nikitin, Filipp and Dunn, Ian and Koes, David Ryan and Isayev, Olexandr},
	month = may,
	year = {2025},
	note = {arXiv:2505.00169 [cs]},
}

@misc{joshi_all-atom_2025,
	title = {All-atom {Diffusion} {Transformers}: {Unified} generative modelling of molecules and materials},
	shorttitle = {All-atom {Diffusion} {Transformers}},
	url = {http://arxiv.org/abs/2503.03965},
	doi = {10.48550/arXiv.2503.03965},
	urldate = {2025-07-24},
	publisher = {arXiv},
	author = {Joshi, Chaitanya K. and Fu, Xiang and Liao, Yi-Lun and Gharakhanyan, Vahe and Miller, Benjamin Kurt and Sriram, Anuroop and Ulissi, Zachary W.},
	month = may,
	year = {2025},
	note = {arXiv:2503.03965 [cs]},
}

@article{wang_improving_2020,
	title = {Improving {Conformer} {Generation} for {Small} {Rings} and {Macrocycles} {Based} on {Distance} {Geometry} and {Experimental} {Torsional}-{Angle} {Preferences}},
	volume = {60},
	issn = {1549-9596},
	url = {https://doi.org/10.1021/acs.jcim.0c00025},
	doi = {10.1021/acs.jcim.0c00025},
	number = {4},
	urldate = {2025-08-13},
	journal = {Journal of Chemical Information and Modeling},
	author = {Wang, Shuzhe and Witek, Jagna and Landrum, Gregory A. and Riniker, Sereina},
	month = apr,
	year = {2020},
	note = {Publisher: American Chemical Society},
	pages = {2044--2058},
}

@article{rappe_uff_1992,
	title = {{UFF}, a full periodic table force field for molecular mechanics and molecular dynamics simulations},
	volume = {114},
	issn = {0002-7863},
	url = {https://doi.org/10.1021/ja00051a040},
	doi = {10.1021/ja00051a040},
	number = {25},
	urldate = {2025-08-13},
	journal = {Journal of the American Chemical Society},
	author = {Rappe, A. K. and Casewit, C. J. and Colwell, K. S. and Goddard, W. A. III and Skiff, W. M.},
	month = dec,
	year = {1992},
	note = {Publisher: American Chemical Society},
	pages = {10024--10035},
}

@misc{dunn_flowmol3_2025,
	title = {{FlowMol3}: {Flow} {Matching} for {3D} {De} {Novo} {Small}-{Molecule} {Generation}},
	shorttitle = {{FlowMol3}},
	url = {http://arxiv.org/abs/2508.12629},
	doi = {10.48550/arXiv.2508.12629},
	urldate = {2025-08-19},
	publisher = {arXiv},
	author = {Dunn, Ian and Koes, David R.},
	month = aug,
	year = {2025},
	note = {arXiv:2508.12629 [cs]},
}

@misc{cremer_flowr_2025,
	title = {{FLOWR}: {Flow} {Matching} for {Structure}-{Aware} {De} {Novo}, {Interaction}- and {Fragment}-{Based} {Ligand} {Generation}},
	shorttitle = {{FLOWR}},
	url = {http://arxiv.org/abs/2504.10564},
	doi = {10.48550/arXiv.2504.10564},
	urldate = {2025-08-19},
	publisher = {arXiv},
	author = {Cremer, Julian and Irwin, Ross and Tibo, Alessandro and Janet, Jon Paul and Olsson, Simon and Clevert, Djork-Arné},
	month = may,
	year = {2025},
	note = {arXiv:2504.10564 [q-bio]
version: 2},
}

@misc{cremer_pilot_2024,
	title = {{PILOT}: {Equivariant} diffusion for pocket conditioned de novo ligand generation with multi-objective guidance via importance sampling},
	shorttitle = {{PILOT}},
	url = {http://arxiv.org/abs/2405.14925},
	doi = {10.48550/arXiv.2405.14925},
	urldate = {2025-08-19},
	publisher = {arXiv},
	author = {Cremer, Julian and Le, Tuan and Noé, Frank and Clevert, Djork-Arné and Schütt, Kristof T.},
	month = may,
	year = {2024},
	note = {arXiv:2405.14925 [q-bio]
version: 1},
}

@article{bilodeau_generative_2022,
	title = {Generative models for molecular discovery: {Recent} advances and challenges},
	volume = {12},
	copyright = {© 2022 The Authors. WIREs Computational Molecular Science published by Wiley Periodicals LLC.},
	issn = {1759-0884},
	shorttitle = {Generative models for molecular discovery},
	url = {https://onlinelibrary.wiley.com/doi/abs/10.1002/wcms.1608},
	doi = {10.1002/wcms.1608},
	language = {en},
	number = {5},
	urldate = {2025-09-12},
	journal = {WIREs Computational Molecular Science},
	author = {Bilodeau, Camille and Jin, Wengong and Jaakkola, Tommi and Barzilay, Regina and Jensen, Klavs F.},
	year = {2022},
	note = {\_eprint: https://wires.onlinelibrary.wiley.com/doi/pdf/10.1002/wcms.1608},
	pages = {e1608},
}

@article{yim_diffusion_2024,
	title = {Diffusion models in protein structure and docking},
	volume = {14},
	copyright = {© 2024 The Authors. WIREs Computational Molecular Science published by Wiley Periodicals LLC.},
	issn = {1759-0884},
	url = {https://onlinelibrary.wiley.com/doi/abs/10.1002/wcms.1711},
	doi = {10.1002/wcms.1711},
	language = {en},
	number = {2},
	urldate = {2025-09-12},
	journal = {WIREs Computational Molecular Science},
	author = {Yim, Jason and Stärk, Hannes and Corso, Gabriele and Jing, Bowen and Barzilay, Regina and Jaakkola, Tommi S.},
	year = {2024},
	note = {\_eprint: https://wires.onlinelibrary.wiley.com/doi/pdf/10.1002/wcms.1711},
	pages = {e1711},
}

@article{koes_pharmer_2011,
	title = {Pharmer: {Efficient} and {Exact} {Pharmacophore} {Search}},
	volume = {51},
	issn = {1549-9596},
	shorttitle = {Pharmer},
	url = {https://doi.org/10.1021/ci200097m},
	doi = {10.1021/ci200097m},
	number = {6},
	urldate = {2025-09-19},
	journal = {Journal of Chemical Information and Modeling},
	author = {Koes, David Ryan and Camacho, Carlos J.},
	month = jun,
	year = {2011},
	note = {Publisher: American Chemical Society},
	pages = {1307--1314},
}

@article{sunseri_pharmit_2016,
	title = {Pharmit: interactive exploration of chemical space},
	volume = {44},
	issn = {0305-1048},
	shorttitle = {Pharmit},
	url = {https://doi.org/10.1093/nar/gkw287},
	doi = {10.1093/nar/gkw287},
	number = {W1},
	urldate = {2025-09-19},
	journal = {Nucleic Acids Research},
	author = {Sunseri, Jocelyn and Koes, David Ryan},
	month = jul,
	year = {2016},
	pages = {W442--W448},
}

@article{francoeur_three-dimensional_2020,
	title = {Three-{Dimensional} {Convolutional} {Neural} {Networks} and a {Cross}-{Docked} {Data} {Set} for {Structure}-{Based} {Drug} {Design}},
	volume = {60},
	issn = {1549-9596},
	url = {https://doi.org/10.1021/acs.jcim.0c00411},
	doi = {10.1021/acs.jcim.0c00411},
	number = {9},
	urldate = {2025-09-19},
	journal = {Journal of Chemical Information and Modeling},
	author = {Francoeur, Paul G. and Masuda, Tomohide and Sunseri, Jocelyn and Jia, Andrew and Iovanisci, Richard B. and Snyder, Ian and Koes, David R.},
	month = sep,
	year = {2020},
	note = {Publisher: American Chemical Society},
	pages = {4200--4215},
}

@misc{peng_pocket2mol_2025,
	title = {{Pocket2Mol}: {Efficient} {Molecular} {Sampling} {Based} on {3D} {Protein} {Pockets}},
	shorttitle = {{Pocket2Mol}},
	url = {http://arxiv.org/abs/2205.07249},
	doi = {10.48550/arXiv.2205.07249},
	urldate = {2025-09-19},
	publisher = {arXiv},
	author = {Peng, Xingang and Luo, Shitong and Guan, Jiaqi and Xie, Qi and Peng, Jian and Ma, Jianzhu},
	month = jul,
	year = {2025},
	note = {arXiv:2205.07249 [cs]},
}

@misc{luo_3d_2022,
	title = {A {3D} {Generative} {Model} for {Structure}-{Based} {Drug} {Design}},
	url = {http://arxiv.org/abs/2203.10446},
	doi = {10.48550/arXiv.2203.10446},
	urldate = {2025-09-19},
	publisher = {arXiv},
	author = {Luo, Shitong and Guan, Jiaqi and Ma, Jianzhu and Peng, Jian},
	month = nov,
	year = {2022},
	note = {arXiv:2203.10446 [q-bio]},
}

@misc{zhang_physdock_2025,
	title = {{PhysDock}: {A} {Physics}-{Guided} {All}-{Atom} {Diffusion} {Model} for {Protein}-{Ligand} {Complex} {Prediction}},
	copyright = {© 2025, Posted by Cold Spring Harbor Laboratory. This pre-print is available under a Creative Commons License (Attribution-NonCommercial 4.0 International), CC BY-NC 4.0, as described at http://creativecommons.org/licenses/by-nc/4.0/},
	shorttitle = {{PhysDock}},
	url = {https://www.biorxiv.org/content/10.1101/2025.04.28.650887v4},
	doi = {10.1101/2025.04.28.650887},
	language = {en},
	urldate = {2025-09-22},
	publisher = {bioRxiv},
	author = {Zhang, Kexin and Ma, Yuanyuan and Yu, Jiale and Luo, Huiting and Lin, Jinyu and Qin, Yifan and Li, Xiangcheng and Jiang, Qian and Bai, Fang and Dou, Jiayi and Zheng, Jie and Yu, Jingyi and Sun, Liping},
	month = jun,
	year = {2025},
	note = {Pages: 2025.04.28.650887
Section: New Results},
}

@article{buttenschoen_posebusters_2024,
	title = {{PoseBusters}: {AI}-based docking methods fail to generate physically valid poses or generalise to novel sequences},
	volume = {15},
	issn = {2041-6539},
	shorttitle = {{PoseBusters}},
	url = {https://pubs.rsc.org/en/content/articlelanding/2024/sc/d3sc04185a},
	doi = {10.1039/D3SC04185A},
	language = {en},
	number = {9},
	urldate = {2025-09-22},
	journal = {Chemical Science},
	author = {Buttenschoen, Martin and Morris, Garrett M. and Deane, Charlotte M.},
	month = feb,
	year = {2024},
	note = {Publisher: The Royal Society of Chemistry},
	pages = {3130--3139},
}

@inproceedings{
igashov_large_2025,
title={Large Drug Discovery Model},
author={Ilia Igashov and Arne Schneuing and Adrian W. Dobbelstein and Irina Morozova and Rebecca Manuela Neeser and Evgenia Elizarova and Philippe Schwaller and Michael M. Bronstein and Bruno Correia},
booktitle={ICLR 2025 Workshop on Generative and Experimental Perspectives for Biomolecular Design},
year={2025},
url={https://openreview.net/forum?id=fbL6nHy2xb}
}

@misc{peng_atom-level_2025,
	title = {Atom-level generative foundation model for molecular interaction with pockets},
	copyright = {© 2025, Posted by Cold Spring Harbor Laboratory. This pre-print is available under a Creative Commons License (Attribution 4.0 International), CC BY 4.0, as described at http://creativecommons.org/licenses/by/4.0/},
	url = {https://www.biorxiv.org/content/10.1101/2024.10.17.618827v2},
	doi = {10.1101/2024.10.17.618827},
	language = {en},
	urldate = {2025-09-22},
	publisher = {bioRxiv},
	author = {Peng, Xingang and Guo, Fenglin and Guo, Ruihan and Sun, Jiayu and Guan, Jiaqi and Jia, Yinjun and Xu, Yan and Huang, Yanwen and Zhang, Muhan and Peng, Jian and Wang, Xinquan and Han, Chuanhui and Wang, Zihua and Ma, Jianzhu},
	month = aug,
	year = {2025},
	note = {ISSN: 2692-8205
Pages: 2024.10.17.618827
Section: New Results},
}

@misc{noauthor_openmmpdbfixer_2025,
	title = {openmm/pdbfixer},
	url = {https://github.com/openmm/pdbfixer},
	urldate = {2025-09-23},
	publisher = {OpenMM},
	month = sep,
	year = {2025},
	note = {original-date: 2013-08-29T22:29:24Z},
}

@misc{morehead_how_2025,
	title = {How to go with the flow: flow matching in bioinformatics and computational biology},
	shorttitle = {How to go with the flow},
	url = {https://www.authorea.com/users/637193/articles/1320146-how-to-go-with-the-flow-flow-matching-in-bioinformatics-and-computational-biology},
	urldate = {2025-09-23},
	author = {Morehead, Alex and Atanackovic, Lazar and Hegde, Akshata and Wang, Yanli and Boadu, Frimpong and Selvaraj, Joel and Tong, Alexander and Krishnapriyan, Aditi and Cheng, Jianlin},
	month = sep,
	year = {2025},
	doi = {10.22541/au.175382408.89466370/v3},
	note = {Publication Title: Authorea
Published: Authorea preprint},
}

@inproceedings{vaswani_attention_2017,
	title = {Attention is {All} you {Need}},
	volume = {30},
	url = {https://papers.nips.cc/paper_files/paper/2017/hash/3f5ee243547dee91fbd053c1c4a845aa-Abstract.html},
	urldate = {2025-09-23},
	booktitle = {Advances in {Neural} {Information} {Processing} {Systems}},
	publisher = {Curran Associates, Inc.},
	author = {Vaswani, Ashish and Shazeer, Noam and Parmar, Niki and Uszkoreit, Jakob and Jones, Llion and Gomez, Aidan N and Kaiser, Ł ukasz and Polosukhin, Illia},
	year = {2017},
}

@article{eberhardt_autodock_2021,
	title = {{AutoDock} {Vina} 1.2.0: {New} {Docking} {Methods}, {Expanded} {Force} {Field}, and {Python} {Bindings}},
	volume = {61},
	issn = {1549-9596},
	shorttitle = {{AutoDock} {Vina} 1.2.0},
	url = {https://doi.org/10.1021/acs.jcim.1c00203},
	doi = {10.1021/acs.jcim.1c00203},
	number = {8},
	urldate = {2025-09-24},
	journal = {Journal of Chemical Information and Modeling},
	author = {Eberhardt, Jerome and Santos-Martins, Diogo and Tillack, Andreas F. and Forli, Stefano},
	month = aug,
	year = {2021},
	note = {Publisher: American Chemical Society},
	pages = {3891--3898},
}

@misc{harris_benchmarking_2023,
	title = {Benchmarking {Generated} {Poses}: {How} {Rational} is {Structure}-based {Drug} {Design} with {Generative} {Models}?},
	shorttitle = {Benchmarking {Generated} {Poses}},
	url = {http://arxiv.org/abs/2308.07413},
	doi = {10.48550/arXiv.2308.07413},
	urldate = {2025-09-24},
	publisher = {arXiv},
	author = {Harris, Charles and Didi, Kieran and Jamasb, Arian R. and Joshi, Chaitanya K. and Mathis, Simon V. and Lio, Pietro and Blundell, Tom},
	month = aug,
	year = {2023},
	note = {arXiv:2308.07413 [q-bio]},
}

@article{bouysset_prolif_2021,
	title = {{ProLIF}: a library to encode molecular interactions as fingerprints},
	volume = {13},
	issn = {1758-2946},
	shorttitle = {{ProLIF}},
	url = {https://doi.org/10.1186/s13321-021-00548-6},
	doi = {10.1186/s13321-021-00548-6},
	number = {1},
	urldate = {2025-09-24},
	journal = {Journal of Cheminformatics},
	author = {Bouysset, Cédric and Fiorucci, Sébastien},
	month = sep,
	year = {2021},
	pages = {72},
}

@misc{morehead_flowdock_2025,
	title = {{FlowDock}: {Geometric} {Flow} {Matching} for {Generative} {Protein}-{Ligand} {Docking} and {Affinity} {Prediction}},
	shorttitle = {{FlowDock}},
	url = {http://arxiv.org/abs/2412.10966},
	doi = {10.48550/arXiv.2412.10966},
	urldate = {2025-09-29},
	publisher = {arXiv},
	author = {Morehead, Alex and Cheng, Jianlin},
	month = mar,
	year = {2025},
	note = {arXiv:2412.10966 [cs]},
}

@misc{wohlwend_boltz-1_2025,
	title = {Boltz-1 {Democratizing} {Biomolecular} {Interaction} {Modeling}},
	copyright = {© 2025, Posted by Cold Spring Harbor Laboratory. This pre-print is available under a Creative Commons License (Attribution 4.0 International), CC BY 4.0, as described at http://creativecommons.org/licenses/by/4.0/},
	url = {https://www.biorxiv.org/content/10.1101/2024.11.19.624167v4},
	doi = {10.1101/2024.11.19.624167},
	language = {en},
	urldate = {2025-09-29},
	publisher = {bioRxiv},
	author = {Wohlwend, Jeremy and Corso, Gabriele and Passaro, Saro and Getz, Noah and Reveiz, Mateo and Leidal, Ken and Swiderski, Wojtek and Atkinson, Liam and Portnoi, Tally and Chinn, Itamar and Silterra, Jacob and Jaakkola, Tommi and Barzilay, Regina},
	month = may,
	year = {2025},
	note = {Pages: 2024.11.19.624167
Section: New Results},
}

@misc{discovery_chai-1_2024,
	title = {Chai-1: {Decoding} the molecular interactions of life},
	copyright = {© 2024, Posted by Cold Spring Harbor Laboratory. This pre-print is available under a Creative Commons License (Attribution-NonCommercial 4.0 International), CC BY-NC 4.0, as described at http://creativecommons.org/licenses/by-nc/4.0/},
	shorttitle = {Chai-1},
	url = {https://www.biorxiv.org/content/10.1101/2024.10.10.615955v2},
	doi = {10.1101/2024.10.10.615955},
	language = {en},
	urldate = {2025-09-29},
	publisher = {bioRxiv},
	author = {Discovery, Chai and Boitreaud, Jacques and Dent, Jack and McPartlon, Matthew and Meier, Joshua and Reis, Vinicius and Rogozhnikov, Alex and Wu, Kevin},
	month = oct,
	year = {2024},
	note = {Pages: 2024.10.10.615955
Section: New Results},
}

@article{qiao_state-specific_2024,
	title = {State-specific protein–ligand complex structure prediction with a multiscale deep generative model},
	volume = {6},
	copyright = {2024 The Author(s), under exclusive licence to Springer Nature Limited},
	issn = {2522-5839},
	url = {https://www.nature.com/articles/s42256-024-00792-z},
	doi = {10.1038/s42256-024-00792-z},
	language = {en},
	number = {2},
	urldate = {2025-09-29},
	journal = {Nature Machine Intelligence},
	author = {Qiao, Zhuoran and Nie, Weili and Vahdat, Arash and Miller, Thomas F. and Anandkumar, Animashree},
	month = feb,
	year = {2024},
	note = {Publisher: Nature Publishing Group},
	pages = {195--208},
}

@article{abramson_accurate_2024,
	title = {Accurate structure prediction of biomolecular interactions with {AlphaFold} 3},
	volume = {630},
	copyright = {2024 The Author(s)},
	issn = {1476-4687},
	url = {https://www.nature.com/articles/s41586-024-07487-w},
	doi = {10.1038/s41586-024-07487-w},
	language = {en},
	number = {8016},
	urldate = {2025-09-29},
	journal = {Nature},
	author = {Abramson, Josh and Adler, Jonas and Dunger, Jack and Evans, Richard and Green, Tim and Pritzel, Alexander and Ronneberger, Olaf and Willmore, Lindsay and Ballard, Andrew J. and Bambrick, Joshua and Bodenstein, Sebastian W. and Evans, David A. and Hung, Chia-Chun and O’Neill, Michael and Reiman, David and Tunyasuvunakool, Kathryn and Wu, Zachary and Žemgulytė, Akvilė and Arvaniti, Eirini and Beattie, Charles and Bertolli, Ottavia and Bridgland, Alex and Cherepanov, Alexey and Congreve, Miles and Cowen-Rivers, Alexander I. and Cowie, Andrew and Figurnov, Michael and Fuchs, Fabian B. and Gladman, Hannah and Jain, Rishub and Khan, Yousuf A. and Low, Caroline M. R. and Perlin, Kuba and Potapenko, Anna and Savy, Pascal and Singh, Sukhdeep and Stecula, Adrian and Thillaisundaram, Ashok and Tong, Catherine and Yakneen, Sergei and Zhong, Ellen D. and Zielinski, Michal and Žídek, Augustin and Bapst, Victor and Kohli, Pushmeet and Jaderberg, Max and Hassabis, Demis and Jumper, John M.},
	month = jun,
	year = {2024},
	note = {Publisher: Nature Publishing Group},
	pages = {493--500},
}

@article{du_machine_2024,
	title = {Machine learning-aided generative molecular design},
	volume = {6},
	copyright = {2024 Springer Nature Limited},
	issn = {2522-5839},
	url = {https://www.nature.com/articles/s42256-024-00843-5},
	doi = {10.1038/s42256-024-00843-5},
	language = {en},
	number = {6},
	urldate = {2025-09-29},
	journal = {Nature Machine Intelligence},
	author = {Du, Yuanqi and Jamasb, Arian R. and Guo, Jeff and Fu, Tianfan and Harris, Charles and Wang, Yingheng and Duan, Chenru and Liò, Pietro and Schwaller, Philippe and Blundell, Tom L.},
	month = jun,
	year = {2024},
	note = {Publisher: Nature Publishing Group},
	pages = {589--604},
}

@misc{corso_diffdock_2023,
	title = {{DiffDock}: {Diffusion} {Steps}, {Twists}, and {Turns} for {Molecular} {Docking}},
	shorttitle = {{DiffDock}},
	url = {http://arxiv.org/abs/2210.01776},
	doi = {10.48550/arXiv.2210.01776},
	urldate = {2025-09-29},
	publisher = {arXiv},
	author = {Corso, Gabriele and Stärk, Hannes and Jing, Bowen and Barzilay, Regina and Jaakkola, Tommi},
	month = feb,
	year = {2023},
	note = {arXiv:2210.01776 [q-bio]},
}

@article{chen_target_2025,
	title = {Target sequence-conditioned design of peptide binders using masked language modeling},
	copyright = {2025 The Author(s)},
	issn = {1546-1696},
	url = {https://www.nature.com/articles/s41587-025-02761-2},
	doi = {10.1038/s41587-025-02761-2},
	language = {en},
	urldate = {2025-09-29},
	journal = {Nature Biotechnology},
	author = {Chen, Leo Tianlai and Quinn, Zachary and Dumas, Madeleine and Peng, Christina and Hong, Lauren and Lopez-Gonzalez, Moises and Mestre, Alexander and Watson, Rio and Vincoff, Sophia and Zhao, Lin and Wu, Jianli and Stavrand, Audrey and Schaepers-Cheu, Mayumi and Wang, Tian Zi and Srijay, Divya and Monticello, Connor and Vure, Pranay and Pulugurta, Rishab and Pertsemlidis, Sarah and Kholina, Kseniia and Goel, Shrey and DeLisa, Matthew P. and Chi, Jen-Tsan Ashley and Truant, Ray and Aguilar, Hector C. and Chatterjee, Pranam},
	month = aug,
	year = {2025},
	note = {Publisher: Nature Publishing Group},
	pages = {1--9},
}

@misc{qin_defog_2025,
	title = {{DeFoG}: {Discrete} {Flow} {Matching} for {Graph} {Generation}},
	shorttitle = {{DeFoG}},
	url = {http://arxiv.org/abs/2410.04263},
	doi = {10.48550/arXiv.2410.04263},
	urldate = {2025-09-29},
	publisher = {arXiv},
	author = {Qin, Yiming and Madeira, Manuel and Thanou, Dorina and Frossard, Pascal},
	month = jun,
	year = {2025},
	note = {arXiv:2410.04263 [cs]},
}

@misc{jing_generating_2025,
	title = {Generating functional and multistate proteins with a multimodal diffusion transformer},
	copyright = {© 2025, Posted by Cold Spring Harbor Laboratory. This pre-print is available under a Creative Commons License (Attribution 4.0 International), CC BY 4.0, as described at http://creativecommons.org/licenses/by/4.0/},
	url = {https://www.biorxiv.org/content/10.1101/2025.09.03.672144v2},
	doi = {10.1101/2025.09.03.672144},
	language = {en},
	urldate = {2025-09-29},
	publisher = {bioRxiv},
	author = {Jing, Bowen and Sappington, Anna and Bafna, Mihir and Shah, Ravi and Tang, Adrina and Krishna, Rohith and Klivans, Adam and Diaz, Daniel J. and Berger, Bonnie},
	month = sep,
	year = {2025},
	note = {ISSN: 2692-8205
Pages: 2025.09.03.672144
Section: New Results},
}

@article{jumper_highly_2021,
	title = {Highly accurate protein structure prediction with {AlphaFold}},
	volume = {596},
	copyright = {2021 The Author(s)},
	issn = {1476-4687},
	url = {https://www.nature.com/articles/s41586-021-03819-2},
	doi = {10.1038/s41586-021-03819-2},
	language = {en},
	number = {7873},
	urldate = {2025-09-29},
	journal = {Nature},
	author = {Jumper, John and Evans, Richard and Pritzel, Alexander and Green, Tim and Figurnov, Michael and Ronneberger, Olaf and Tunyasuvunakool, Kathryn and Bates, Russ and Žídek, Augustin and Potapenko, Anna and Bridgland, Alex and Meyer, Clemens and Kohl, Simon A. A. and Ballard, Andrew J. and Cowie, Andrew and Romera-Paredes, Bernardino and Nikolov, Stanislav and Jain, Rishub and Adler, Jonas and Back, Trevor and Petersen, Stig and Reiman, David and Clancy, Ellen and Zielinski, Michal and Steinegger, Martin and Pacholska, Michalina and Berghammer, Tamas and Bodenstein, Sebastian and Silver, David and Vinyals, Oriol and Senior, Andrew W. and Kavukcuoglu, Koray and Kohli, Pushmeet and Hassabis, Demis},
	month = aug,
	year = {2021},
	note = {Publisher: Nature Publishing Group},
	pages = {583--589},
}

@article{cao_surfdock_2025,
	title = {{SurfDock} is a surface-informed diffusion generative model for reliable and accurate protein–ligand complex prediction},
	volume = {22},
	copyright = {2024 The Author(s), under exclusive licence to Springer Nature America, Inc.},
	issn = {1548-7105},
	url = {https://www.nature.com/articles/s41592-024-02516-y},
	doi = {10.1038/s41592-024-02516-y},
	language = {en},
	number = {2},
	urldate = {2025-09-30},
	journal = {Nature Methods},
	author = {Cao, Duanhua and Chen, Mingan and Zhang, Runze and Wang, Zhaokun and Huang, Manlin and Yu, Jie and Jiang, Xinyu and Fan, Zhehuan and Zhang, Wei and Zhou, Hao and Li, Xutong and Fu, Zunyun and Zhang, Sulin and Zheng, Mingyue},
	month = feb,
	year = {2025},
	note = {Publisher: Nature Publishing Group},
	pages = {310--322},
}

@misc{anishchenko_modeling_2025,
	title = {Modeling protein-small molecule conformational ensembles with {PLACER}},
	copyright = {© 2025, Posted by Cold Spring Harbor Laboratory. This pre-print is available under a Creative Commons License (Attribution 4.0 International), CC BY 4.0, as described at http://creativecommons.org/licenses/by/4.0/},
	url = {https://www.biorxiv.org/content/10.1101/2024.09.25.614868v2},
	doi = {10.1101/2024.09.25.614868},
	language = {en},
	urldate = {2025-10-01},
	publisher = {bioRxiv},
	author = {Anishchenko, Ivan and Kipnis, Yakov and Kalvet, Indrek and Zhou, Guangfeng and Krishna, Rohith and Pellock, Samuel J. and Lauko, Anna and Lee, Gyu Rie and An, Linna and Dauparas, Justas and DiMaio, Frank and Baker, David},
	month = sep,
	year = {2025},
	note = {ISSN: 2692-8205
Pages: 2024.09.25.614868
Section: New Results},
}

@article{mcnutt_gnina_2025,
	title = {{GNINA} 1.3: the next increment in molecular docking with deep learning},
	volume = {17},
	issn = {1758-2946},
	shorttitle = {{GNINA} 1.3},
	url = {https://doi.org/10.1186/s13321-025-00973-x},
	doi = {10.1186/s13321-025-00973-x},
	number = {1},
	urldate = {2025-10-01},
	journal = {Journal of Cheminformatics},
	author = {McNutt, Andrew T. and Li, Yanjing and Meli, Rocco and Aggarwal, Rishal and Koes, David Ryan},
	month = mar,
	year = {2025},
	pages = {28},
}

@article{mcnutt2023conformer,
  title={Conformer generation for structure-based drug design: How many and how good?},
  author={McNutt, Andrew T and Bisiriyu, Fatimah and Song, Sophia and Vyas, Ananya and Hutchison, Geoffrey R and Koes, David Ryan},
  journal={Journal of Chemical Information and Modeling},
  volume={63},
  number={21},
  pages={6598--6607},
  year={2023},
  publisher={ACS Publications}
}

@article {plinder,
	author = {Durairaj, Janani and Adeshina, Yusuf and Cao, Zhonglin and Zhang, Xuejin and Oleinikovas, Vladas and Duignan, Thomas and McClure, Zachary and Robin, Xavier and Studer, Gabriel and Kovtun, Daniel and Rossi, Emanuele and Zhou, Guoqing and Veccham, Srimukh and Isert, Clemens and Peng, Yuxing and Sundareson, Prabindh and Akdel, Mehmet and Corso, Gabriele and St{\"a}rk, Hannes and Tauriello, Gerardo and Carpenter, Zachary and Bronstein, Michael and Kucukbenli, Emine and Schwede, Torsten and Naef, Luca},
	title = {PLINDER: The protein-ligand interactions dataset and evaluation resource},
	elocation-id = {2024.07.17.603955},
	year = {2024},
	doi = {10.1101/2024.07.17.603955},
	publisher = {Cold Spring Harbor Laboratory},
	URL = {https://www.biorxiv.org/content/early/2024/07/19/2024.07.17.603955.1},
	eprint = {https://www.biorxiv.org/content/early/2024/07/19/2024.07.17.603955.1.full.pdf},
	journal = {bioRxiv}
}



\setcounter{section}{0}
\setcounter{figure}{0}
\setcounter{table}{0}
\setcounter{equation}{0}
\renewcommand{\thesection}{A\arabic{section}}
\renewcommand{\thefigure}{A\arabic{figure}}
\renewcommand{\thetable}{A\arabic{table}}
\renewcommand{\theequation}{A\arabic{equation}}
\renewcommand{\thesubsection}{A\arabic{section}.\arabic{subsection}}

\clearpage
\thispagestyle{plain}
\begin{center}
  {\Huge\bfseries Appendix for OMTRA}\\[0.5em]
  \rule{\linewidth}{0.6pt}
\end{center}

\addcontentsline{toc}{section}{Appendix}

\makeatletter
\newcommand{\appendixtableofcontents}{%
  \section*{Appendix Contents}%
  \@starttoc{atoc}%
}

\newcommand{\startappendixtoc}{%
  \begingroup
  \let\app@oldsection\section
  \let\app@oldsubsection\subsection
  \renewcommand{\section}[1]{%
    \app@oldsection{##1}%
    \addcontentsline{atoc}{section}{\protect\numberline{\thesection}##1}}
  \renewcommand{\subsection}[1]{%
    \app@oldsubsection{##1}%
    \addcontentsline{atoc}{subsection}{\protect\numberline{\thesubsection}##1}}
}
\newcommand{\stopappendixtoc}{\endgroup}
\makeatother

\appendixtableofcontents
\clearpage

\startappendixtoc



\section{Extended Results}

\subsection{Performance on No-Protein Tasks}
In Table \ref{tab:no-prot} we provide the performance of OMTRA on tasks that do not involve protein structures. \% PB-Valid is the fraction of molecules passing all PoseBusters~\cite{buttenschoen_posebusters_2024} checks.  \%~Valid Bond Lengths indicates molecules with physically reasonable bond distances, as determined by PoseBusters.  \%~Chirality Preserved reflects the fraction of tetrahedral stereocenters retaining their configuration (relevant for conformer generation).  
\%~Pharm Match is the fraction of molecules in which all ground-truth pharmacophore centers are matched by at least one generated functional group of the same type within 1~Å.  

Overall, unconditional generation achieves near-perfect PB-validity and bond lengths but low stereochemistry retention (17.2\%), while pharmacophore conditioning improves chirality preservation (17.2\% $\rightarrow$ 32.8\%) with a slight decline in structural validity.

\begin{table}[H]
\centering
\caption{Selected performance metrics for 500 samples per task. \label{tab:no-prot}
}
\resizebox{\textwidth}{!}{%
\begin{tabular}{@{}l c c c c@{}}
\toprule
Task & \% PB-Valid & \% Valid Bond Lengths & \% Chirality Preserved & \% Pharm Match \\
\midrule
Unconditional De Novo Design & 99.4 & 100.0 & -- & -- \\
Unconditional Conformer Generation & 100.0 & 100.0 & 17.2 & -- \\
Pharm-Conditioned De Novo Design & 97.1 & 99.6 & -- & 98.7 \\
Pharm-Conditioned Conformer Generation & 98.0 & 98.7 & 32.8 & 99.3 \\
\bottomrule
\end{tabular}}
\end{table}

\subsection{Trajectories}
\begin{figure}[H]
    \centering
    \includegraphics[width=\linewidth]{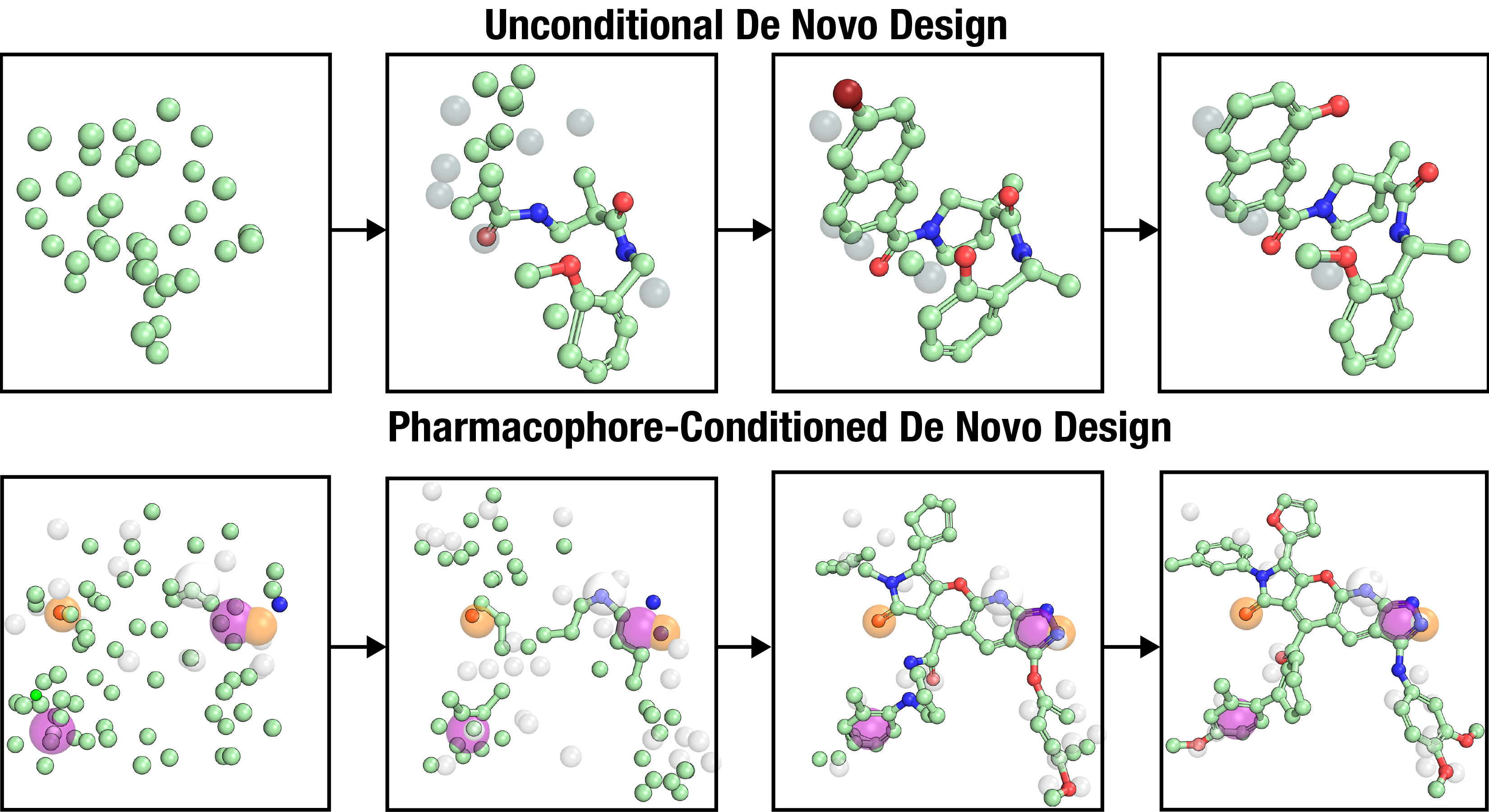}
    \caption{Trajectories of Ligand Generation Using OMTRA. Each row demonstrates the stepwise evolution of ligand structures across OMTRA’s sampling trajectory. The top row illustrates unconditional de novo design, where OMTRA generates ligand structures without guiding constraints. The bottom row illustrates pharmacophore-conditioned de novo design, where OMTRA generates ligand structures guided by pre-defined pharmacophores.}
    \label{fig:noprot-traj}
\end{figure}

\begin{figure}[H]
    \centering
    \includegraphics[width=\linewidth]{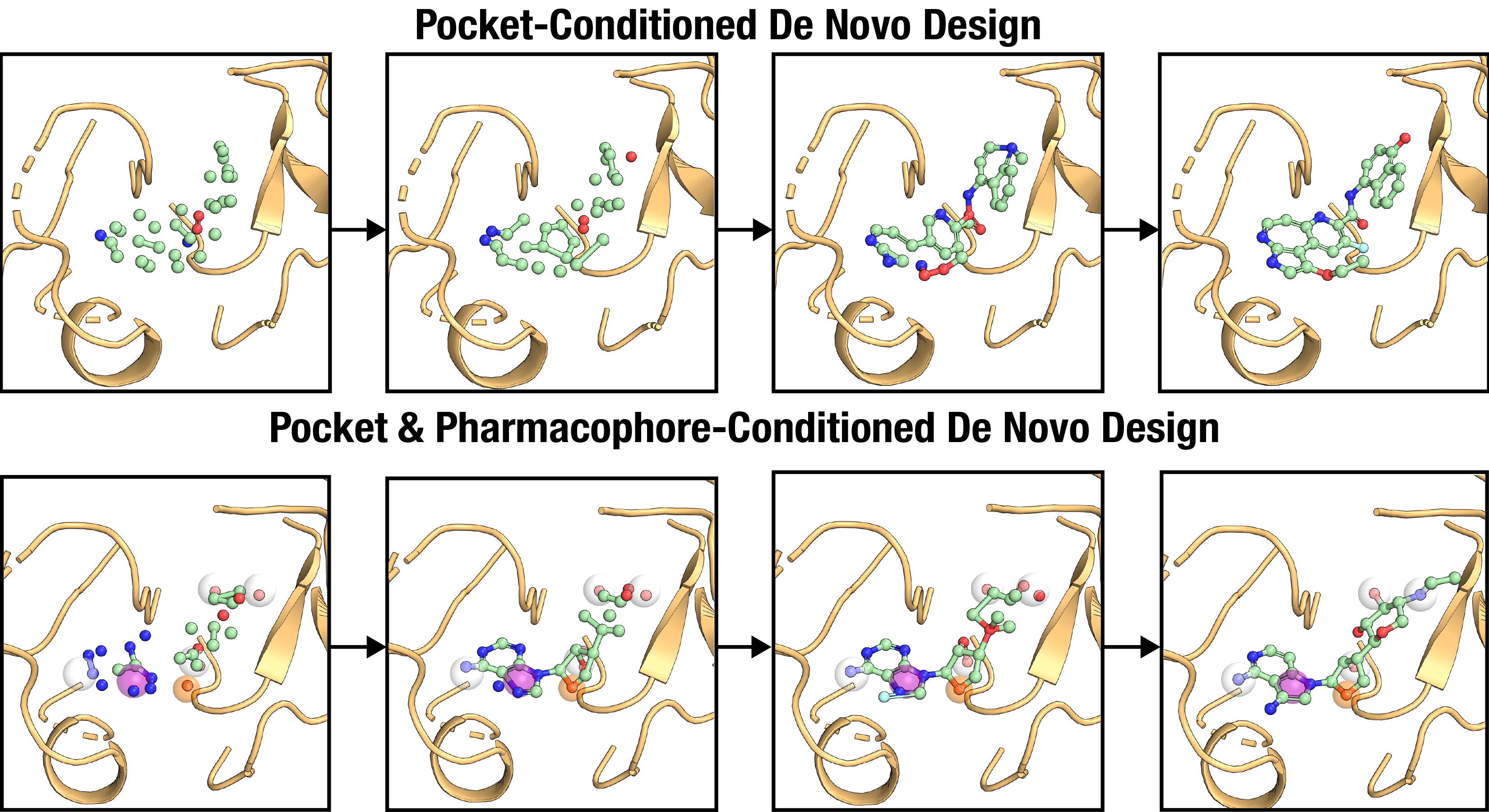}
    \caption{Trajectories of Ligand Generation Inside Pocket Using OMTRA. Each row demonstrates the stepwise evolution of ligand structures across OMTRA's sampling trajectory. The top row illustrates pocket-conditioned de novo design, where OMTRA generates ligand structures guided by the receptor pocket. The bottom row illustrates pocket and pharmacophore-conditioned de novo design, where OMTRA generates ligand structures guided by both the receptor pocket and pre-defined pharmacophores.}
    \label{fig:prot-traj}
\end{figure}

\subsection{PoseBusters Checks}

\begin{figure}[H]
    \centering
    \includegraphics[width=\linewidth]{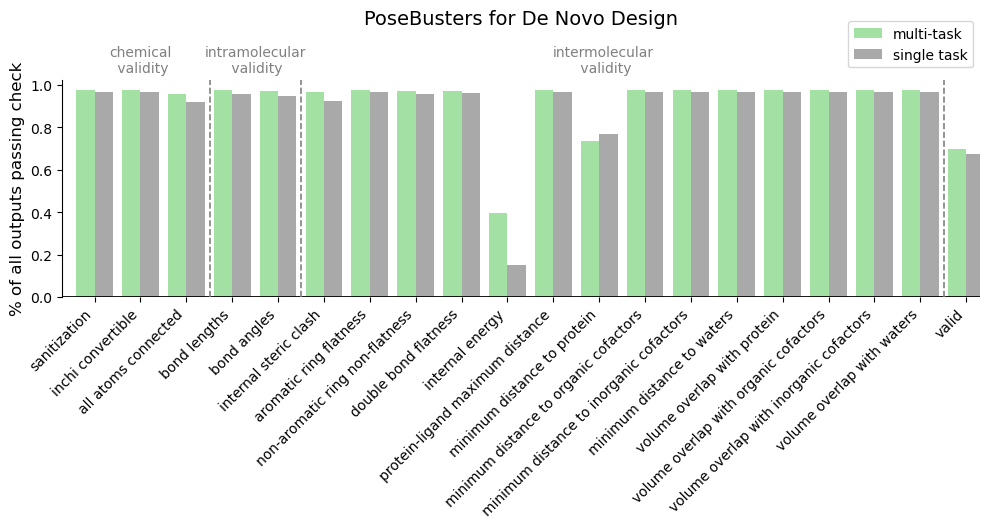}
    \vspace{1em}
    \includegraphics[width=\linewidth]{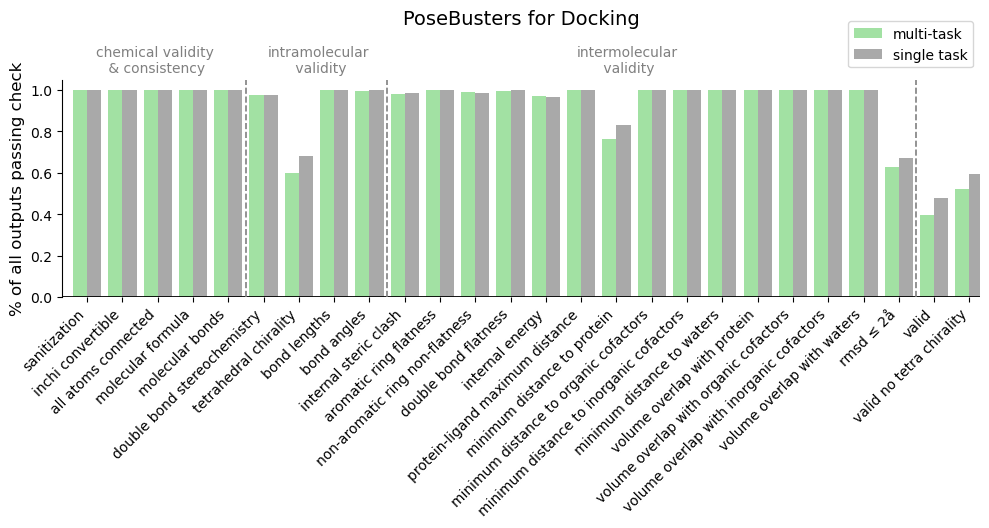}
    \caption{\textbf{PoseBusters Checks}. Comparison of multi-task versus single task OMTRA models on PoseBusters checks for \textit{de novo} design (Top) and docking (Bottom). All models were pretrained on unconditional ligand generation tasks. Metrics report the fraction of passing ligands, calculated from 100 samples per pocket across 100 proteins in the Plinder test split.}
    \label{fig:posebusters_combined}
\end{figure}

We report the pass rates of OMTRA-docked molecules on individual PoseBusters tests in Figure \ref{fig:posebusters_combined}. Importantly here we report pass rates on the \textit{average} docked pose. By contrast, in the main paper, we sample poses, rank them, and then compute metrics on the top-N poses for each pocket. The mean value of metrics on docked poses without ranking is substantially worse. Drawing multiple samples and ranking them therefore remains an essential step in obtaining high-quality predictions. This is consistent with other state-of-the-art transport-based generative models.

\subsection{Examples of Top De Novo Ligands}
\begin{figure}[H]
    \centering
    \includegraphics[width=\linewidth]{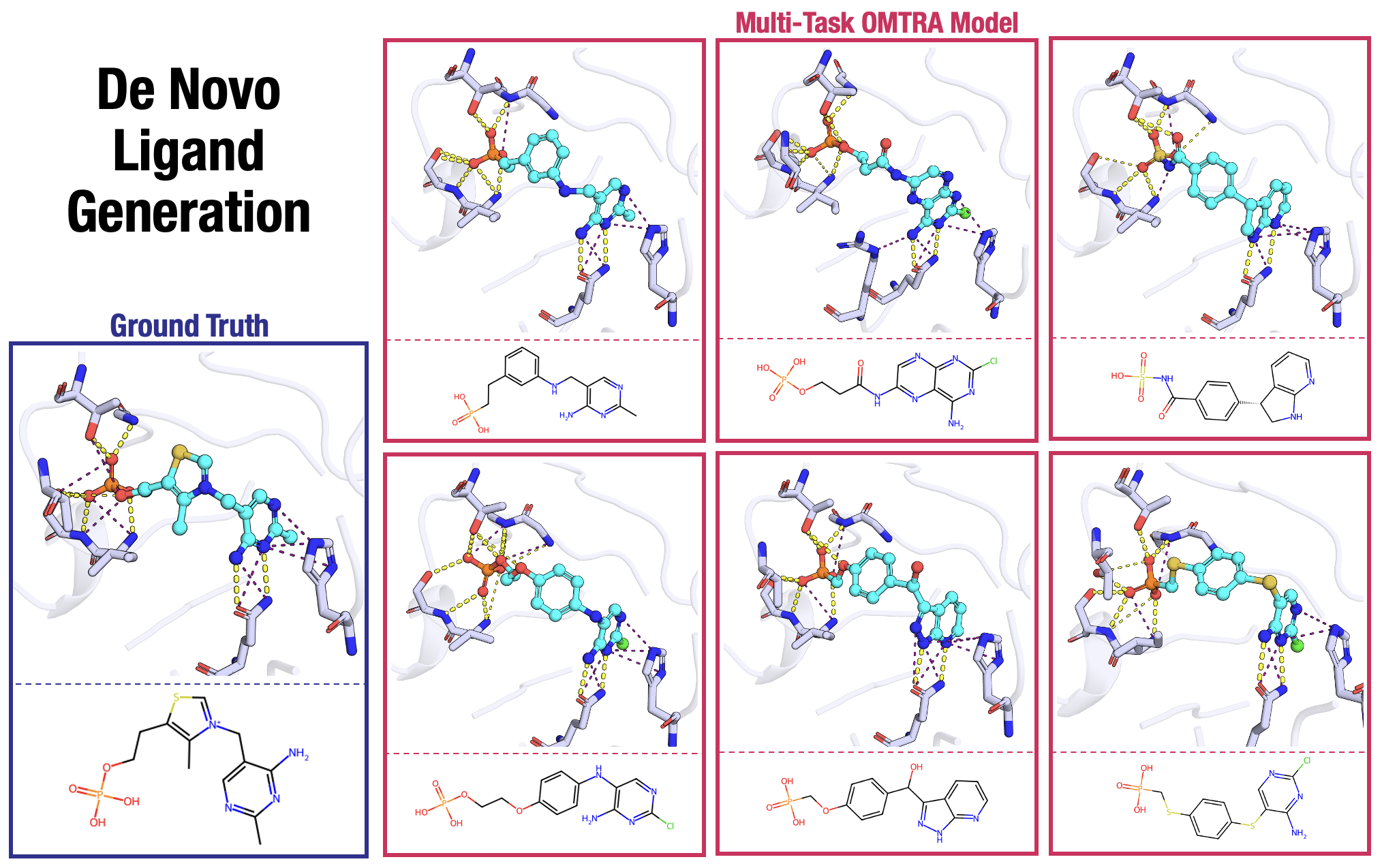}
    \caption{\textbf{Top \textit{De Novo} Ligands Generated by OMTRA Multi-Task Model}. Ligand generation was conditioned on the binding pocket of thiamin phosphate synthase (PDB: 1G4S). Protein-ligand contacts are shown as dashed lines. The ground truth ligand, thiamin phosphate (CCD: TPS), is shown in the left panel. \textit{De novo} generated samples (Right) for this target have an interaction recovery rate of 85.7-100\%.}
    \label{fig:denovo_lig_samples.png}
\end{figure}

\subsection{Effects of Pharmacophore Conditioning} \label{ap:pharm-effect}

\begin{figure}[H]
    \centering
    \includegraphics[width=0.999\linewidth]{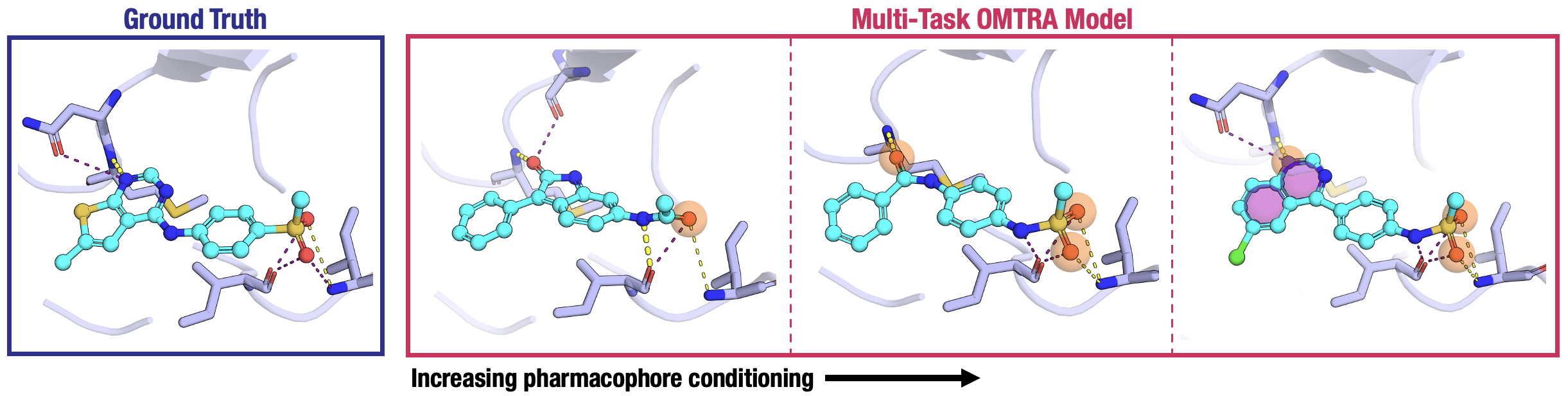}
    \caption{\textbf{Pharmacophore-Guided \textit{De Novo} Design}: The left-panel shows a ground-truth complex of a PI5P4K inhibitor (PDB 8BQ4) from the Plinder test set. The three panels on the right side show \textit{de novo} designed ligands that were conditioned on both the pocket structure and varying numbers of pharmacophore centers. Pharmacophore centers are shown as transparent spheres. Polar contacts are shown with dashed lines.}
    \label{fig:pharm_conditioning}
\end{figure}

\begin{figure}[H]
    \centering
    \includegraphics[width=0.95\linewidth]{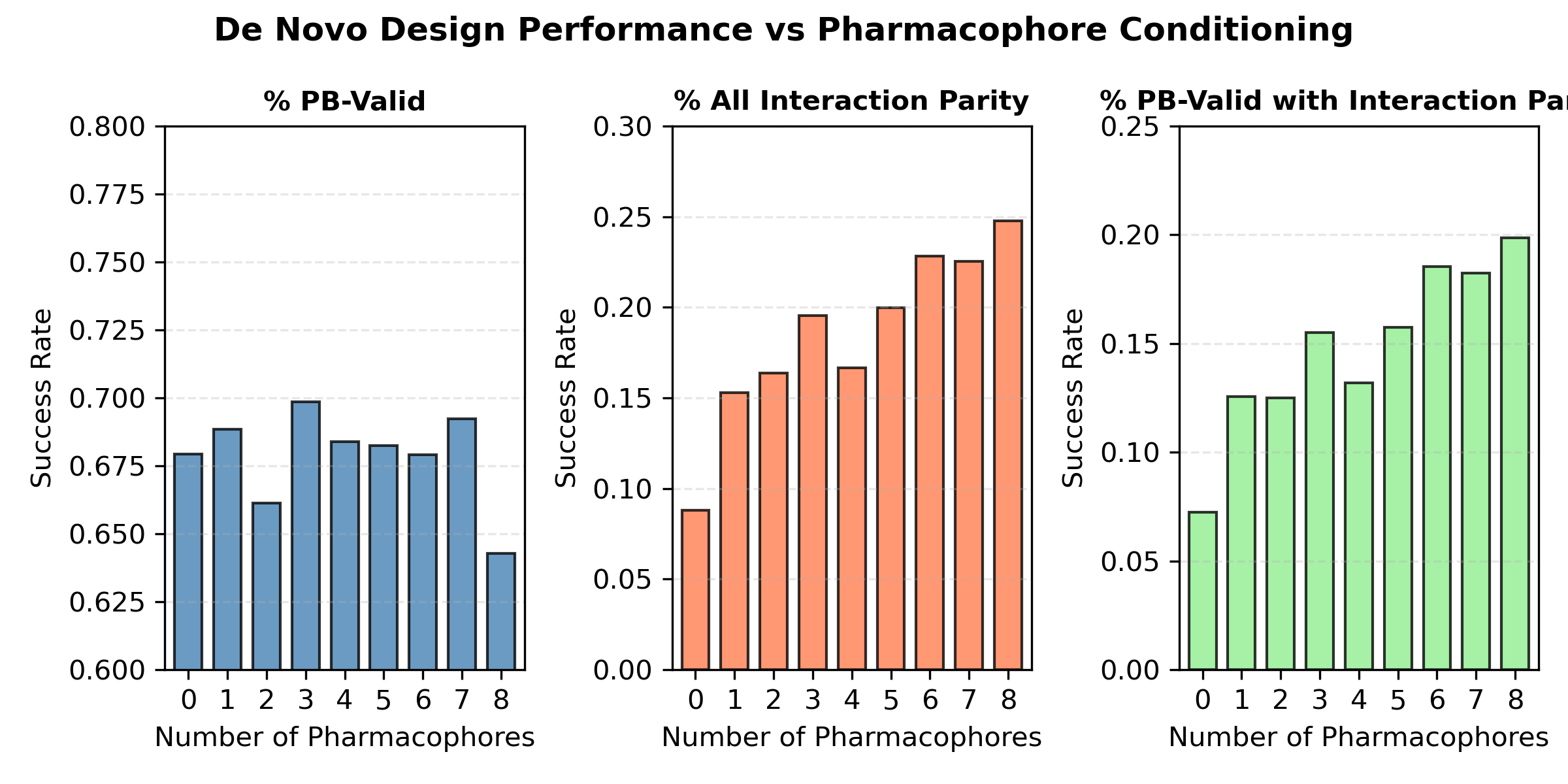}
    \caption{\textbf{\textit{De Novo} Design Quality as a Function of Pharmacophore Conditioning}. Each bar plot shows a metric for the quality of designed ligands. The x-axis is the number of pharmacophore centers that were used as conditioning information when designing a ligand. The model used here was trained and evaluated on the Plinder train and test sets, respectively. Approximately 40,000 samples were generated for this figure.}
    \label{fig:denovo_pharm_effect}
\end{figure}

\begin{figure}[H]
    \centering
    \includegraphics[width=0.95\linewidth]{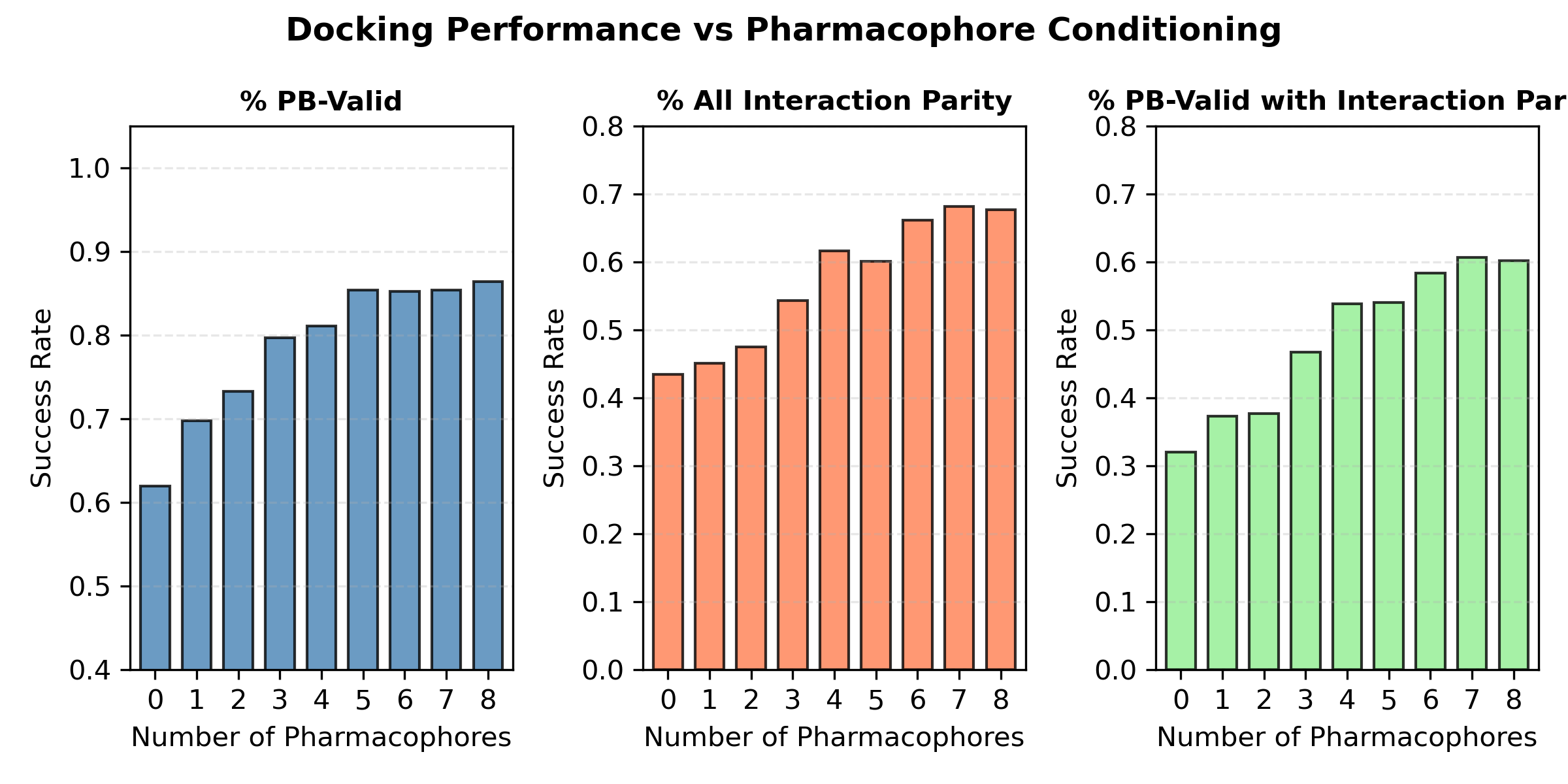}
    \caption{\textbf{Docking Success as a Function of Pharmacophore Conditioning}. Each bar plot shows a metric for the quality of docked ligands. The x-axis is the number of pharmacophore centers that were used as conditioning information when docking a ligand. These values are the mean over all samples; no ranking or filtering has been applied to the samples drawn. The model used here was trained and evaluated on the Plinder train and test sets, respectively. Approximately 40,000 samples were generated for this figure.}
    \label{fig:dock_pharm_effect}
\end{figure}

\section{Related Work}

There have been two concurrent works, to our knowledge, that have proposed a variant of multi-task models for structure based drug design. These are the Large Drug Discovery Model (LDDM) \cite{igashov_large_2025} and PocketXMol \cite{peng_atom-level_2025}. Both models perform tasks related to small-molecule SBDD. To our knowledge, neither work has presented ablation experiments to measure the effects of multi-task training vs training several single-task models. Additionally, to our knowledge, neither of these works supports pharmacophores as a modality.

Several works have proposed variations of multi-task models in other domains. \citet{campbell_generative_2024} proposed a joint protein sequence and backbone model that can perform arbitrary tasks from altering these modalities; forward-folding, inverse-folding, and co-folding. \citet{huguet_sequence-augmented_2024} proposed a protein structure generative model that could either perform unconditional backbone sampling or sequence-conditioned backbone sampling. \citet{reidenbach_applications_2025} trained two variants of the same model for both unconditional conformer generation and unconditional de novo design, but not one model for both tasks.

\citet{cremer_pilot_2024} and \citet{cremer_flowr_2025} have proposed unconditional pre-training for pocket-conditioned de novo generative models.

\section{Experiments}


Our experiments address three themes: the role of ligand-only pretraining, the impact of multi-task training, and the utility of pharmacophore conditioning.

\paragraph{Ligand-Only Pretraining.}
We evaluate single-task OMTRA models for pocket-conditioned \textit{de novo} design and molecular docking, each with and without Pharmit pretraining. Pretraining mixes conformer generation (50\%) and unconditional design (50\%), trained for \(\sim\)5 days on 2 L40 GPUs. This yields four OMTRA variants.

\paragraph{Multi-Task Training.}
OMTRA is jointly trained on \textit{de novo} design and docking, initialized from the Pharmit-pretrained models, and compared against Pharmit-pretrained single-task counterparts.

\paragraph{Protein + Pharmacophore Conditioning.}
We extend OMTRA to simultaneously incorporate protein and pharmacophore context. This enables pocket-conditioned \textit{de novo} design where pharmacophore constraints bias generation toward reproducing specific interactions, and docking where pharmacophore-guided soft constraints influence pose prediction. These tasks, to our knowledge, are unexplored in prior work and demonstrate how OMTRA can guide \textit{de novo} design and docking with interpretable, human-provided priors.

\section{Problem Setup}
\label{ap:problem-setup}

\paragraph{Graphs.}
Each biomolecular system is represented as a heterogeneous graph
\[
G = (V, E),
\]
with node types
\[
\mathcal{T}_V = \{\text{ligand atom (L)}, \text{protein atom (P)}, \text{pharmacophore (Ph)}, \text{NPNDE (O)}\},
\]
and edge types
\[
\mathcal{T}_E = \{\text{ligand--ligand}, \text{ligand--protein}, \text{protein--protein}, \ldots\}.
\]
Ligand atoms form a complete subgraph $G[L]$. Explicit hydrogens are omitted unless they are essential for defining the molecule identity as determiend by rdkit; this is described in Section \ref{ap:ligand-processing}. NPNDE stands for ``Non-Protein Non-Designable Entity''; these are molecules that exist in a system that are not the target ligand nor the protein, but are biologically relevant. These can include ions, co-factors, and post-translational modifications. NPNDEs are described in detail in Section \ref{ap:npndes}.

\paragraph{Modalities.}
A \emph{modality} is any property of the system to be modeled. The modality set is
\[
\mathcal{M} = \mathcal{M}_{\text{disc}} \cup \mathcal{M}_{\text{cont}},
\]
with $\mathcal{M}_{\text{disc}}$ discrete (e.g., atom types, bond orders) and $\mathcal{M}_{\text{cont}}$ continuous (e.g., 3D coordinates). Each modality $m \in \mathcal{M}$ is attached to a node type $t \in \mathcal{T}_V$ or edge type $t \in \mathcal{T}_E$.  
Multi-modal flow matching treats these jointly: each $m \in \mathcal{M}$ admits a conditional probability path $\{p_t^m\}_{t \in [0,1]}$, which may be continuous or discrete. Sampling is factorized across modalities, but training occurs in a single model with multiple heads, minimizing a weighted sum of modality-specific losses \eqref{eq:multi-modal-loss}.

\paragraph{Tasks.}
A \emph{task} $\tau$ is defined by a partition of modalities,
\[
\mathcal{M} = \mathcal{M}_{\text{gen}}^\tau \cup \mathcal{M}_{\text{cond}}^\tau \cup \mathcal{M}_{\text{abs}}^\tau,
\]
where generated modalities $\mathcal{M}_{\text{gen}}^\tau$ are modeled via flow matching, conditioned modalities $\mathcal{M}_{\text{cond}}^\tau$ are fixed conditioning information for the task, and absent modalities $\mathcal{M}_{\text{abs}}^\tau$ are not modeled in the system.  
For each $m \in \mathcal{M}_{\text{gen}}^\tau$, the task specifies:
\begin{itemize}
    \item a prior distribution $p_0^m$ (source distribution),  
    \item a coupling $\pi^m$ with the data distribution,  
    \item a conditional probability path $p_t^m(x|x_0,x_1)$ defining the flow.  
\end{itemize}
This ties the task formalization directly to multi-modal flow matching: generated modalities evolve under their own flows, while conditioning ensures compatibility across modalities.

\paragraph{OMTRA Instantiation.}
An instantiation of OMTRA is parameterized by a set of tasks
\[
\mathcal{T} = \{\tau_1, \dots, \tau_K\}.
\]
If $|\mathcal{T}| > 1$, parameters are shared across tasks, yielding a multi-task model. Each $\tau \in \mathcal{T}$ corresponds to a distinct generative problem—e.g., unconditional \textit{de novo} ligand design, pocket-conditioned design, or docking—yet all are unified by the same multi-modal flow matching framework. A comprehensive list of tasks and modalities currently supported by OMTRA are provided in Appendix \ref{ap:tasks-and-modalities}.

\section{Data Processing and Representation}

We use multiple datasets but apply consistent data processing and define a consistent molecular representation, so that OMTRA can be trained across all of them. Here we describe our representation of biomolecular systems as seen by the OMTRA model. 

\subsection{Ligands} \label{ap:ligand-processing}

Molecules were filtered if they contained elements outside of a pre-determined supported set (C, H, N, O, F, P, S, Cl, Br, I, and B). Molecules are sanitized by rdkit. Following prior work~\cite{dunn_flowmol3_2025,nikitin_geom-drugs_2025}, all molecules are kekulized. Molecules that fail to be sanitized and kekulilzed are excluded from the dataset. Atoms were stripped of hydrogens via rdkit; notably this function will preserve some hydrogens in cases that they are essential for preserving molecule identity. 
We store bond orders between all pairs of ligand atoms as discrete edge features. The additional essential features to encode ligand identity are the element of every atom and its formal charge. \citet{le_navigating_2023} demonstrated that incorporating other atom-level descriptions such as hybridization as additional modalities in the generative process can improve performance. Towards this end, for every ligand atom we use rdkit to compute the number of implicit hydrogens, aromaticity, hybridization state, ring membership, and chirality for each atom. However, rather than defining seven discrete modalities on ligand atoms, we condense them all down to one discrete modality for atom type; we refer to this technique as ``condensed atom typing''. Specifically, we search our datasets for all unique observed tuples of (atom element, formal charge, number of implicit hydrogens, aromaticity, hybridization state, ring membership, chirality), and use this set of observed values to define our ``vocabulary'' for assigning ligand atoms to discrete types. Thus we only model one discrete modality on every ligand atom. The advantage of condensed atom typing is that (i) atom types are more informative and correlated with 3D geometry and local topological structure than if we used the strictly necessary modalities of atom element + formal charge and (ii) there are fewer modalities to jointly sample, leaving less room for error. We find that training models for conformer generation and de novo design with condensed atom typing gives better performance than either 1. defining all the discrete ligand atom features as separate modalities or 2. only using atom type and formal charge as the ligand atom typing modalities. 



\subsection{Pharmacophores} \label{ap:pharms}

Pharmacophores are abstract representations of the molecular features responsible for key interactions with biological targets. In our framework, pharmacophore features are identified using SMARTS-based pattern matching via RDKit, with a predefined dictionary of SMARTS patterns corresponding to commonly recognized feature types.

The pharmacophore features extracted include:
\begin{itemize}
  \item \textbf{Aromatic rings:} five- and six-membered aromatic ring systems.
  \item \textbf{Hydrogen bond donors:} nitrogen, oxygen, and sulfur atoms bonded to one or more hydrogen atoms.
  \item \textbf{Hydrogen bond acceptors:} nitrogen and oxygen atoms capable of accepting hydrogen bonds, excluding those involved in conjugation or resonance.
  \item \textbf{Hydrophobic groups:} nonpolar regions such as alkyl chains and hydrophobic ring systems, including certain sulfur-containing groups.
  \item \textbf{Halogens:} fluorine, chlorine, bromine, and iodine atoms bonded to carbon atoms.
  \item \textbf{Positively charged groups:} atoms or functional groups carrying a formal positive charge.
  \item \textbf{Negatively charged groups:} atoms or functional groups carrying a formal negative charge.
\end{itemize}

For each matched SMARTS pattern, the 3D coordinates of the corresponding atoms are obtained from the ligand’s conformer, and the centroid of these atoms defines the pharmacophore feature’s spatial location. From a modeling perspective, each pharmacophore feature is a point in Cartesian space with an associated discrete type. Thus, pharmacophore features are represented as nodes in graphs, and two modalities (position, type) are associated with pharmacophore nodes.

In the context of a protein binding pocket with a reference ligand, only pharmacophore features of the reference ligand within a cutoff distance of compatible pocket atoms (i.e. those likely to interact) are retained, and a random subset of 1–8 features is selected to be retained as conditioning information. For tasks without a protein pocket, sub-sampling is performed from the full set of available pharmacophore features. This enables flexible pharmacophore-conditioned generation in both pocket-aware and pocket-free settings, where the model is guided either by potential protein–ligand interactions or by the spatial distribution of ligand functional groups.

\subsection{Protein Pockets} \label{ap:protein-repo}

Water, hydrogens, and deuterium atoms were stripped from protein structures. Binding pockets were extracted by selecting residues with at least one atom within 8\AA\ of a ligand atom. 

Since our graph neural network operates on all-atom representations of molecular structure, we employ a few techniques to enable OMTRA to better reason about the molecular structure beyond just a ball of atoms. On every protein atom node there is an associated discrete modality for the chemical element. There is also a discrete modality for the atom name, which uniquely identifies the atom of each canonical amino acid. This enhances the model's ability to distinguish backbone carbons from sidechain carbons, for example. 

We also include a discrete modality on protein atom nodes indicating the type of residue that the atom belongs to. We support non-canonical amino acid types while still keeping the discrete vocabulary small by mapping non-canonical amino acids to their closest canonical amino acid. We use a pre-defined substitution map obtained from pdbfixer \cite{noauthor_openmmpdbfixer_2025}.

Finally, we include a residue positional encoding that is the sinusoidal positional encoding from \citet{vaswani_attention_2017} applied to the residue index, broadcasted out to each atom belonging to that residue. This feature was critical for enabling flexible-protein tasks but not for the tasks presented / evaluated in this paper.

\subsection{NPNDEs} \label{ap:npndes}

Protein complexes in the PLINDER and Crossdocked dataset may contain multiple small, non-protein molecules. We distinguish between ligands (designable elements of interest) and non-designable, non-protein entities (NPNDE). All small molecules were retained in context, but models were not trained to generate non-designable elements. These were defined as: experimental artifacts, cofactors, ions, covalently bonded saccharides, molecules with > 120 heavy atoms or unresolved atoms, molecules forming more than one covalent bond to the protein, or molecules interacting with fewer than one protein residue.

NPNDE representation is same as ligands but the NPNDE graph is not fully connected. We also do not perform condensed atom typing, nor do we store extra features required for condensed atom typing.

\subsection{Linked Apo Structures}

For PLINDER entries linked to unbound or AlphaFold-predicted structures, we first superposed the linked structure onto the bound complex prior to pocket extraction. We store 3 copies of the Plinder dataset; all plinder systems, plinder systems with exerimental apo linked structure, and plinder systems with AlphaFold-predicted linked structures.




\section{Datasets} \label{ap:datasets}

\subsection{Pharmit} \label{ap:pharmit}

Pharmit~\cite{sunseri_pharmit_2016} is an open-source web application for rapid screening of pharmacophores against databases of conformers using the pharmer algorithm~\cite{koes_pharmer_2011}. As such, since its implementation, its creator has been downloading smiles strings of publicly available molecule databases and using rdkit ETKDG\cite{wang_improving_2020} + UFF minimization~\cite{rappe_uff_1992} to generate 3D conformers \cite{mcnutt2023conformer}, so as to enable pharmacophore screening against these databases. 

We converted every molecule on the Pharmit production server (that passed filtering criteria) into regular numpy arrays stored in Zarr. We storm the atom type, formal charge, chirality, hybridization, aromaticity, and the number of implicit hydrogens on every atom, as well as the bonding topology. 

The Pharmit Dataset is comprised of following major public and commercial libraries: ChEMBL34, ChemDiv, ChemSpace, Enamine, MCULE, MCULE-ULTIMATE, MolPort, NCI Open Chemical Repository, PubChem, LabNetwork, and ZINC. This aggregation provides broad coverage of drug-like chemical space across academic, government, and vendor-curated collections. Molecules were filtered if they contained elements outside of a pre-determined supported set and if they could not be sanitized and kekulized by rdkit, resulting in a dataset of more than 500 million unique molecules. We retain only 1 conformer per molecule, the lowest energy conformer obtained by our conformer generation method. 

\subsection{Plinder}

The PLINDER dataset comprises over 400,000 annotated protein-ligand interaction systems curated from the Protein Data Bank \cite{plinder}. A subset of systems also contain paired unbound and/or predicted structures. PLINDER additionally provides training/evaluation splits based on their novel approach to minimize leakage by leveraging similarity metrics at the level of the protein, pocket, interaction, and ligand.


To ensure high structural quality, we applied multiple filters to the PLINDER dataset to ensure OMTRA was only trained on useful examples. Systems were excluded if they had crystallographic resolution worse than 3.5 Å, $R > 0.40$, $R_\text{free} > 0.45$, or $|R - R_\text{free}| > 0.075$. Additionally, we leveraged the PoseBusters metadata to remove systems in which ligands exhibited a volume overlap greater than 0.075 with the protein or cofactors. 

We computed the fraction of ligand atoms coming within 4\AA\ of a protein pocket atom for each system; systems having fewer than 35\% of ligand atoms in contact with protein atoms were also filtered from the final dataset. Systems may fail to meet this threshold due to interactions with membranes or symmetry mates, which are not modeled by OMTRA.

\subsection{Crossdocked} 

The Crossdocked dataset \cite{francoeur_three-dimensional_2020}, derived from experimentally determined protein-ligand complexes in PDBbind \cite{liu_pdbbind_2017}, comprises of over 22.5 million protein-ligand poses generated by systematically docking ligands into both cognate and noncognate receptors, grouped by binding site similarity \cite{francoeur_three-dimensional_2020}. Through docking 13,780 unique ligands to multiple similar binding pockets, the dataset captures realistic variability in protein conformation and ligand orientation, enabling the training and evaluation of models across multiple structure-based tasks.

To reduce overlap between similar protein structures and chemotypes in the training and test sets, the original Crossdocked2020 paper adopted clustered cross-validation splits for both PDBBind and Crossdocked data. For the PDBBind datasets, clusters were defined by grouping receptors with >50\% sequence identity, or with >40\% sequence identity and >90\% ligand similarity, as computed using RDKit fingerprints. Under this scheme, highly similar ligands are only assigned to different clusters if the corresponding receptors share less than 40\% sequence identity, reducing information leakage across splits \cite{francoeur_three-dimensional_2020}. For the CrossDocked2020 dataset, clustering was based on Pocketome-defined pockets, further grouped using the ProBiS structural alignment algorithm with a z-score threshold of 3.5, enabling cross-validation across structurally similar binding sites \cite{francoeur_three-dimensional_2020}.

However, in this work we adopt a different split scheme, introduced by Luo et al. (2022). These splits were generated by clustering data at 30\% sequence identity, and randomly drawing 100,000 protein-ligand pairs for training, and 100 proteins from remaining clusters for testing \cite{luo_3d_2022,peng_pocket2mol_2025}.

Although these splits are less rigorous than those of the original Crossdocked authors and likely contain information leakage, they are widely reused. As such, we use these splits to enable standardized evaluation across models.

\subsection{Posebusters Benchmark Set}
The PoseBusters Benchmark set is a dataset of 428 diverse protein-ligand complexes originally curated by \citet{buttenschoen_posebusters_2024}. This is a moderately difficult docking benchmark comprising 428 diverse protein-ligand complexes. We use this dataset to compare OMTRA's re-docking performance to existing models. We sample 40 docked poses for each system and rank OMTRA-docked poses using the Autodock Vina~\cite{eberhardt_autodock_2021} scoring function. Notably in our evaluations of the PoseBusters benchmark set, we did not take care to remove overlapping examples from the Plinder training set, and as such our estimates of docking performance may be optimistic. 

\section{Flow Matching Formulation} \label{ap:flow-matching}


Our flow matching formulation closely follows that of \citet{dunn_flowmol3_2025}. Continuous and discrete modalities, and the multi-modal flow matching formulations are nearly identical; there are just more modalities supported under OMTRA.

\subsection{Introduction}

Flow matching \cite{tong_improving_2023,albergo_stochastic_2023,lipman_flow_2023,liu_flow_2022} prescribes a method to interpolate between two distributions $q_{\text{source}}$ and $q_{\text{target}}$ by modeling a family of intermediate distributions $\{p_t\}_{t\in[0,1]}$ with $p_0 = q_{\text{source}}$ and $p_1 = q_{\text{target}}$. Samples evolve along conditional probability paths defined by a coupling $p(x_0,x_1)$ between initial and final states. Given a conditioning variable $z=(x_0,x_1)$, the marginal path satisfies

\begin{equation}
    p_t(x) = \mathbb{E}_{p(z)}\!\left[ p_t(x|z) \right]
\end{equation}

with the property that $x_t \sim p_t(x|x_0,x_1)$ can be simulated without iterative sampling.

\paragraph{Continuous Flow Matching}  
For continuous variables $x \in \mathbb{R}^d$, the marginal process is sampled by a vector field $u_t(x)=\tfrac{dx}{dt}$. The vector field may be learned directly, or indirectly by predicting the conditional expectation of the final state $\mathbb{E}[x_1|x_t]$. Training then reduces to denoising: sampling $x_t \sim p_t(x|x_0,x_1)$ and minimizing error between the network prediction $\hat{x}_1(x_t)$ and the true $x_1$:

\begin{equation} \label{eq:denoiseloss}
\mathcal{L}_{\text{EFM}} = \mathbb{E}_{t,p(x_0,x_1),p_t(x_t|x_0,x_1)} 
\Big[ (1-t)^{-2}\,\|\hat{x}_1(x_t) - x_1\|^2 \Big]
\end{equation}

\paragraph{Discrete Flow Matching}  
For discrete data $x \in \mathcal{V}^N$ (sequences of tokens from a vocabulary $\mathcal{V}$), trajectories are governed by continuous-time Markov chains (CTMCs). Each token remains in its current state until it jumps to a new one. In masked discrete flow matching \cite{campbell_generative_2024}, the prior is a sequence of mask tokens, and the marginal CTMC is parameterized by a denoiser that estimates unmasked states in partially-masked sequences. Training uses a cross-entropy objective:

\begin{equation} \label{eq:dfm-loss}
\mathcal{L}_{\text{CE}} = \mathbb{E}_{t,p(x_0,x_1),p_t(x_t|x_0,x_1)} 
\left[ -\sum_{i=1}^N \log p_{1|t}^\theta(x_1^i \mid x_t)\,\delta_M(x_t^i) \right]
\end{equation}


\paragraph{Multi-Modal Flow Matching}  
Data structures often involve multiple modalities $\mathcal{M}$, e.g., continuous positions and discrete types in molecular systems. Multi-modal flow matching \cite{dunn_flowmol3_2025,campbell_generative_2024} extends the framework by factorizing conditional paths across modalities. A single neural network with multiple heads predicts outputs for each modality, with the overall loss being a weighted sum over modalities:

\begin{equation} \label{eq:multi-modal-loss}
    \mathcal{L} = \sum_{m\in\mathcal{M}} \lambda_m \mathcal{L}_m
\end{equation}

For OMTRA, we use the denoising loss \eqref{eq:denoiseloss} and cross-entropy loss \eqref{eq:dfm-loss} for continuous and discrete modalities, respectively. 

\subsection{Design Choices for Continuous Flow Matching}

\paragraph{Prior and Coupling }For continuous data $x \in \mathbb{R}^d$ (atom positions), our prior is a Gaussian distribution with mean equal to the center-of-mass of the ground-truth ligand and variance of 2.573\AA\, which is the mean variance of ligand atomic coordinates in the data distribution. For conformer generation or docking, where the ligand identity is conditioning information, the coupling distribution for atomic coordinates is the independent coupling. For \textit{de novo} ligand generation, the coupling distribution involves a distance-minimizing permutation of (initial,final) atom pair assignments as proposed in \citet{dunn_flowmol3_2025}.

\paragraph{Conditional Probability Path} The conditional probability path is generated implicitly by the geometry distortion interpolant proposed by \citet{dunn_flowmol3_2025}:

\begin{equation} \label{eq:distortion-path}
    X_t = (1 - t)\,X_0 + t\,X_1 + \mathbb{I}[t \geq t_{\mathrm{distort}}]\,\bigl(M \odot \varepsilon\bigr),
\end{equation}

where $\mathbb{I}$ is the indicator function, $\odot$ is the Hadamard product, $M \in [0,1]^N$ is a binary mask over atoms having the property $M_i \sim \mathrm{Bernoulli}(p_{\mathrm{distort}})$, and $\epsilon \in \mathbb{R}^{N\times 3}$ is a per-atom displacement having the property $\epsilon_i \sim \mathcal{N}(0, \sigma_{\mathrm{distort}} I_3)$. Geometry distortion is controlled by three hyperparameters that are set to $p_{\mathrm{distort}} = 0.2$, $t_{\mathrm{distort}} = 0.5$, and $\sigma_{\mathrm{distort}} = 0.5$.

\subsection{Design Choices for Discrete Flow Matching}

For all discrete modalities generated by OMTRA, we employ discrete flow matching (DFM) as developed by \citet{campbell_generative_2024} and \citet{gat_discrete_2024}. Specifically this is masked discrete flow matching where the prior is a sequence of a special mask state. At inference time, individual tokens can jump between mask tokens, unmasked tokens, and back to masked tokens. The rate at which remasking happens can be controlled by an inference-time hyperparameter $\eta$. The training objective is as described by \eqref{eq:dfm-loss}; our model produces, for each node/edge, logits over possible discrete states at $t=1$ given the current state at time $t$. For details on the conditional probability paths and sampling techniques, we refer the reader to \citet{dunn_flowmol3_2025}.

\section{Supported Tasks and Modalities} \label{ap:tasks-and-modalities}
OMTRA supports a variety of modalities, each representing a distinct type of input features for a particular task. Modality examples include ligand and protein structure, identity, pharmacophore types, and so forth. A complete list of modalities used in OMTRA is outlined in Table \ref{tab:modal}. These modalities are grouped by conceptual function and are used as building blocks across multiple generative tasks.

OMTRA also supports a variety of tasks, each representing a different generative or predictive objective. Tasks specify which entities are present and determine how modalities are used, either as constraints or generated outputs. Task examples include de novo ligand design, pocket and pharmacophore conditioned de novo design, and rigid docking. A complete list of currently supported and upcoming tasks is outlined in Table \ref{tab:modal} and Table \ref{tab:unsupported-task} respectively. 

\subsection{Supported Modalities}
\begin{table}[H]
\centering
\setlength{\tabcolsep}{8pt}
\caption{OMTRA Supported Modalities. The modality \texttt{Ligand Condensed Features} includes implicit hydrogens, aromaticity, hybridization, in ring, and chirality, all represented as categorical variables. Condensed atom typing is described in detail in Section \ref{ap:ligand-processing}.}
\label{tab:modal}
\begin{tabular}{llll} 
\toprule
\textbf{Modality} & \textbf{Group} & \textbf{Data Type} & \textbf{Graph Entity} \\
\midrule

Ligand Atom Positions & Ligand Structure & Continuous & Node \\
\midrule

Ligand Condensed Atom Features & Ligand Identity & Discrete & Node \\
Ligand Bond Orders & Ligand Identity & Discrete & Edge \\
\midrule

Pharmacophore Positions & Pharmacophore & Continuous & Node \\
Pharmacophore Types & Pharmacophore & Discrete & Node \\
\midrule

Protein Atom Positions & Protein Structure & Continuous & Node \\
NPNDE Positions & Protein Structure & Continuous & Node \\
\midrule

Protein Residue Names & Protein Identity & Discrete & Node \\
Protein Atom Elements & Protein Identity & Discrete & Node \\
Protein Atom Names & Protein Identity & Discrete & Node \\
NPNDE Atom Elements & Protein Identity & Discrete & Node \\
NPNDE Charges & Protein Identity & Discrete & Node \\
NPNDE Bond Order & Protein Identity & Discrete & Edge \\
\bottomrule
\end{tabular}
\end{table}

\clearpage
\subsection{Supported Tasks}
\begin{table}[H]
\centering
\label{tab: supported task}
\caption{OMTRA Supported Tasks}
\newcolumntype{L}[1]{>{\raggedright\arraybackslash}p{#1}}
\begin{tabular}{L{3cm} L{3cm} L{4cm} L{4cm}}
\toprule
\textbf{Task} & \textbf{Entities Present} & \textbf{Modality Groups Fixed} & \textbf{Modality Groups Generated} \\

\midrule
Unconditional De Novo Ligand Design & Ligand & None & Ligand Identity, Ligand Structure \\
\midrule
Ligand Conformer Design & Ligand & Ligand Identity & Ligand Structure \\

\midrule
Rigid Docking & Ligand \& Protein & Ligand Identity, Protein Identity, Protein Structure & Ligand Structure \\
\midrule
Pocket-Conditioned De Novo Ligand Design & Ligand \& Protein & Protein Identity, Protein Structure & Ligand Identity, Ligand Structure \\

\midrule
Pharmacophore-Conditioned De Novo Ligand Design & Ligand \& Pharmacophore & Pharmacophore & Ligand Identity, Ligand Structure \\
\midrule
Pharmacophore-Conditioned Ligand Conformer Design & Ligand \& Pharmacophore & Ligand Identity, Pharmacophore & Ligand Structure \\

\midrule
Pharmacophore-Conditioned Rigid Docking & Ligand, Protein, \& Pharmacophore & Protein Identity, Protein Structure, Ligand Identity, Pharmacophore & Ligand Structure \\
\midrule
Pocket and Pharmacophore-Conditioned De Novo Ligand Design & Ligand, Protein, \& Pharmacophore & Protein Identity, Protein Structure, Pharmacophore & Ligand Identity, Ligand Structure \\
\bottomrule
\end{tabular}
\end{table}
\clearpage
\subsection{Upcoming Tasks}
\begin{table}[H]
\centering
\caption{OMTRA Upcoming Tasks}
\label{tab:unsupported-task}
\newcolumntype{L}[1]{>{\raggedright\arraybackslash}p{#1}}
\begin{tabular}{L{3cm} L{3cm} L{4cm} L{4cm}}
\toprule
\textbf{Task} & \textbf{Entities Present} & \textbf{Modality Groups Fixed} & \textbf{Modality Groups Generated} \\

\midrule
Predicted Apo, Flexible Protein De Novo Ligand Design & Ligand \& Protein & Protein Identity & Ligand Identity, Ligand Structure, Protein Structure \\

\midrule
Flexible Protein Docking & Ligand \& Protein & Ligand Identity, Protein Identity & Ligand Structure, Protein Structure \\
\midrule
Flexible Protein and De Novo Ligand Design & Ligand \& Protein & Protein Identity & Protein Structure, Ligand Identity, Ligand Structure \\
\midrule
Experimental Apo-Conditioned, Flexible Protein Docking & Ligand \& Protein & Ligand Identity, Protein Identity & Ligand Structure, Protein Structure \\
\midrule
Predicted Apo-Conditioned, Flexible Protein Docking & Ligand \& Protein & Ligand Identity, Protein Identity & Ligand Structure, Protein Structure \\

\midrule
Unconditional De Novo Ligand and Pharmacophore Design & Ligand \& Pharmacophore & None & Ligand Identity, Ligand Structure, Pharmacophore \\

\midrule
Flexible Protein and De Novo Pharmacophore Design & Protein \& Pharmacophore & Protein Identity & Protein Structure, Pharmacophore \\
\midrule
Experimental Apo-Conditioned, Flexible Protein and De Novo Pharmacophore Design & Protein \& Pharmacophore & Protein Identity & Protein Structure, Pharmacophore \\

\midrule
De Novo Protein, Ligand, and Pharmacophore Design & Ligand, Protein, \& Pharmacophore & Protein Identity & Protein Structure, Ligand Identity, Ligand Structure, Pharmacophore \\
\bottomrule
\end{tabular}
\end{table}

\section{Architecture} \label{ap:arch}

Our architecture is essentialy the simplest possible extension of FlowMol3 \cite{dunn_flowmol3_2025} to heterogeneous graphs with type-specific message passing and feature updates. Nodes $i \in V$ carry Cartesian positions $x_i \in \mathbb{R}^3$, scalar features $s_i \in \mathbb{R}^{d_s}$, and vector features $v_i \in \mathbb{R}^{d_v \times 3}$. Edges $(i,j) \in E$ are typed; ligand--ligand and other--other edges additionally carry scalar features $e_{ij} \in \mathbb{R}^{d_e}$. Operations on $(x_i, v_i)$ are SE(3)-equivariant; operations on $(s_i, e_{ij})$ are SE(3)-invariant. Node vectors $v_i$ are first-order geometric vectors relative to $x_i$. SE(3)-equivariant operations are enabled by a custom variant of GVPs~\cite{jing_equivariant_2021} introduced in \citet{dunn_flowmol3_2025}.

\subsection{Initial Feature Embeddings}

All nodes in the system are initialized with scalar embeddings of a fixed latent size (256). Additionally, ligand-ligand and NPNDE-NPNDE edges also have initial scalar edge embeddings.

Scalar node embeddings are initialized by the node modalities defined on that node type that are not positions. For ligand atoms, these are the condensed atom types. We use a standard embedding lookup table (pytorch's nn.Embedding module) to map each discrete type to a learnable fixed-length vector. The only continuous modality on node features that is not a position is the residue positional embedding described in Section \ref{ap:protein-repo}. The modality embeddings are concatenated together along with a sinusoidal time embedding and a learnable task embedding. These features are passed through a scalar embedding MLP that is unique to each node type to generate intital scalar node embeddings.

Ligand-ligand and NPNDE-NPNDE edges are simply embedded with an nn.Embedding lookup of the bond order on that edge.

\subsection{Graph Convolutions}

\paragraph{Message Passing.}
Messages are generated per edge type $r \in \mathcal{T}_E$:
\begin{equation} \label{eq:generic-msg-fn}
m_{i \leftarrow j}^r = \phi_r(s_i, s_j, v_i, v_j, e_{ij}, x_i - x_j),
\end{equation}

where $\phi_r$ is an edge-type–specific function. Each node aggregates all incoming messages from all edge types:
\begin{equation}
    M_i = \sum_{r \in \mathcal{T}_E} \sum_{j \in \mathcal{N}_r(i)} m_{i \leftarrow j}^r.
\end{equation}

\subsection{Node and Edge Feature Updates}

Each node type $\alpha \in \mathcal{T}_V$ has its own update function $\psi_\alpha$:
\begin{equation}
    (s_i, v_i) \;\;\mapsto\;\; \psi_{\alpha(i)}(s_i, v_i, M_i),
\end{equation}

which updates only scalar and vector features. Node positions $x_i$ are not inputs to $\psi_\alpha$ and are updated separately. We defer the reader to \citet{dunn_flowmol3_2025} for the precise node-update step equation.

Graph convolutions (message passing + node feature updates) are interleaved with:  
\begin{itemize}
    \item \textbf{Node position updates:} node-wise GVPs operating on $(s_i, v_i, x_i)$ to produce new $x_i$.  
    \item \textbf{Edge feature updates:} edge-wise MLPs that take $(e_{ij}, s_i, s_j, x_i, x_j)$ to produce new $e_{ij}$.  
\end{itemize}

\subsection{Block Structure}
Two graph convolutions followed by position and edge updates form one \emph{ConvolutionBlock}; OMTRA stacks 4 such blocks.

\subsection{Model Outputs and Multi-Tasking}
For node positions being generated, final $x_i$ are taken directly from the last \emph{ConvolutionBlock} output as these blocks update node positions internally. The final latent node/edge features are mapped to categorical logits by shallow MLP heads. If a graph lacks nodes or edges of a given type, the corresponding $\phi_r$ or $\psi_\alpha$ is simply skipped. Node-position updates and categorical logit heads are applied only when the relevant modality is being generated under the current task.

\subsection{Message Generation}

Here we describe the specific form our message-generating functions \eqref{eq:generic-msg-fn}. The message generating functions are chains of GVPs
~\cite{jing_generating_2025}. GVPs accept and return a tuple of scalar and vector features. Therefore, scalar and vector messages $m_{i \to j}^{(s)}$ and $m_{i \to j}^{(v)}$ are generated by the message-generating function $\phi_M$, which is 3 GVPs chained together.

\begin{equation}
    \label{gvp_mij}
    m_{i \to j}^{(s)}, m_{i \to j}^{(v)} = \phi_M \left(
    \left[ s_i^{(l)} : e_{ij}^{(l)} : d_{ij}^{(l)} \right]
    , \left[v_i : \frac{x_i^{(l)} - x_j^{(l)}}{d_{ij}^{(l)}} \right]\right) 
\end{equation}

where $:$ denotes concatenation and $d_{ij}$ is the distance between nodes $i$ and $j$ at molecule update block $l$. In practice, we replace all instances of $d_{ij}$ with a radial basis embedding of that distance before passing through GVPs or MLPs.

\section{Protein-Task Evaluation Pipeline}

\subsection{De Novo Evaluation Methods} \label{ap:denovo-eval}

We benchmark OMTRA against four models for pocket-conditioned, de novo ligand design: DrugFlow, a flow matching model; DiffSBDD and TargetDiff, two diffusion models; and Pocket2Mol, an autoregressive model \cite{peng_pocket2mol_2025,schneuing_multi-domain_2025,guan_3d_2023,schneuing_structure-based_2024}. We evaluated the models on the Crossdocked test split published by Luo et al. \cite{luo_crossdocked_external_splits}. For each protein pocket, we sample 100 ligands.

The PoseBusters suite is widely used to evaluate the chemical and geometric plausibility of small-molecule ligands and their interactions with protein binding pockets \cite{buttenschoen_posebusters_2024}. In the context of pocket-conditioned \textit{de novo} design, PoseBusters checks whether the ligand can be sanitized by RDKit, has reasonable geometry, is not fragmented, has reasonable internal ligand energy, has minimal clashes, and has reasonable proximity to other molecules in the system such as the protein and co-factors. We define the metric \%PB-Valid as the fraction of ligands that pass all PoseBusters checks. Ligands that cannot be sanitized are automatically marked as not PB-Valid, and other individual tests implemented by posebusters are not performed.

We use Gnina~\cite{mcnutt_Gnina_2025} to obtain scores from Autodock Vina~\cite{eberhardt_autodock_2021} (this amounts to running Gnina with the --score-only option). 

The strain energy of the generated ligand measures how much internal energy is stored in the ligand, with a lower strain generally corresponding to more favorable binding interactions with the pocket. It is computed by comparing the difference in energy before and after energy minimization. Relaxation and energy computation were performed by PoseCheck using the Universal Force Field (UFF)~\cite{rappe_uff_1992}. 

Interaction fingerprints were extracted using PoseCheck~\cite{harris_benchmarking_2023} which uses ProLIF~\cite{bouysset_prolif_2021} to identify different protein-ligand interactions based on smarts patterns and geometries \cite{harris_benchmarking_2023, bouysset_prolif_2021}. PoseCheck reports, for each interaction type, the number of interactions occurring in a protein-ligand complex divided by the number of ligand atoms. We use these normalized interaction counts to compute the metrics ``interaction parity'' and ``interaction recovery''. The interactions types we use for these analyses are hydrogen bond donors, hydrogen bond acceptors, and hydrophobic interactions. 

Interaction parity measures whether the number of interactions occurring between a generated ligand and a target protein are greater than or equal to that of the ground-truth ligand. Interaction parity is a binary value, a pass/fail test, performed on each generated ligand. For each interaction type, we check if the normalized interaction count is greater than or equal to the normalized interaction count for the ground-truth ligand. Interaction parity is ``true'' for a single ligand if the normalized interaction counts are greater than that of the ground-truth ligand for all interaction types. The ``\% Interaction Parity'' reported in the main body are the fraction of sampled ligands that have interaction parity. 

``Interaction recovery'' is defined as when when the generated ligand reproduces the same types of interactions with the same residue as the ground truth ligand. The interaction recovery rate is the fraction of interactions made by the ground truth ligand that are recovered by the \textit{de novo} generated ligand. These interaction-based metrics give insight into whether the models reproduce the specific interactions observed in known binding ligands.


For tasks that include pharmacophores, we report the additional metric "\% Pharm Matches", which is the fraction of pharmacophores in the ground truth ligand that are matched by a pharmacophore in the generated ligand. A match requires both the correct pharmacophore identity as well as the right location, defined as being within 1Å of the ground truth pharmacophore.

\subsection{Docking Evaluation Methods} \label{ap:docking-eval}
  

PoseBusters was also used to evaluate the docked poses. In its docking evaluation mode, PoseBusters includes the checks described for \textit{de novo} design evaluation together with additional analyses that use the ground truth ligand conformation. These assess whether the molecular formula, bonds, tetrahedral chirality, and double bond stereochemistry of the generated ligand match the ground truth, as well as whether the RMSD between the generated and the ground truth conformation is <2\AA\ . We report how frequently docked poses are <2\AA\ from the ground truth. We also report \%PB-Valid which is the fraction of docked poses that are within <2\AA\ RMSD from the ground-truth and pass all of the PoseBusters physical plausibility checks.

To be consistent with prior works~\cite{buttenschoen_posebusters_2024,abramson_accurate_2024,zhang_physdock_2025,cao_surfdock_2025}, we report docking metrics on the ``Top N'' sampled poses. This requires sampling multiple docked poses, and then ranking them by some metric.  After ranking poses amd selecting the top-N, we compute: (i) the fraction of systems with at least one pose in the top N whose RMSD is <2\AA\ and (ii) the fraction of systems with at least one pose among the top-N whose RMSD is <2\AA\ and is PB-Valid. We report these metrics for N=1 and N=5. 

Each of the baseline docking methods evaluated has its own ranking method; the deep-learning based docking methods have their own confidence modules for ranking, and conventional docking methods use the same scoring function to rank poses as was used to guide MCMC sampling of poses. OMTRA poses are ranked by a scoring function that is the Autodock Vina score augmented with an additional penalty for violation of chirality. This practice is consistent with AlphaFold3~\cite{abramson_accurate_2024} which ranks poses by a composite score including the confidence and penalties for clashes and chirality violations. 

For tasks with pharmacophore conditioning, we again report the metric "\% Pharm Matches". Here, since ligand identity is fixed, the metric simply measures the degree to which the pharmacophores of the generated ligand conformation are aligned with those of the ground truth ligand. We also report interaction recovery as described in the preceding section.

\stopappendixtoc

\end{document}